\documentclass[journal]{IEEEtran}

\usepackage{times}
\usepackage{graphicx}
\usepackage{multirow}
\usepackage{booktabs}
\usepackage{algorithm}
\usepackage{multirow}
\usepackage{xcolor}
\usepackage{algorithmicx}
\usepackage{algpseudocode}
\usepackage{mathrsfs}
\usepackage{amsmath}
\usepackage{amssymb}
\usepackage{mathrsfs}
\DeclareMathOperator*{\argmin}{argmin}

\usepackage{fancyhdr}

\begin{document}
\title{Learning from Naturalistic Driving Data for Human-like Autonomous Highway Driving}

\author{Donghao~Xu,
		Zhezhang~Ding,
		Xu~He,
        Huijing~Zhao,
        Mathieu~Moze,
        Fran\c{c}ois~Aioun,
        and~Franck~Guillemard
\thanks{This work is partially supported by Groupe PSA's OpenLab program (Multimodal Perception and Reasoning for Intelligent Vehicles) and the NSFC Grants 61573027. D. Xu, Z. Ding, X. He and H. Zhao are with the Key Lab of
Machine Perception (MOE), Peking University, Beijing, China; M. Moze, F. Aioun and F. Guillemard are with Groupe PSA, Velizy, France. Contact: H. Zhao, zhaohj@cis.pku.edu.cn.}
}

\markboth{IEEE TRANSACTIONS ON INTELLIGENT TRANSPORTATION SYSTEMS,~Vol.~?, No.~?, ??~????}%
{Shell \MakeLowercase{\textit{et al.}}: IEEE TRANSACTIONS ON INTELLIGENT TRANSPORTATION SYSTEMS}

\maketitle

\begin{abstract}
Driving in a human-like manner is important for an autonomous vehicle to be a smart and predictable traffic participant. To achieve this goal, parameters of the motion planning module should be carefully tuned, which needs great effort and expert knowledge. In this study, a method of learning cost parameters of a motion planner from naturalistic driving data is proposed. The learning is achieved by encouraging the selected trajectory to approximate the human driving trajectory under the same traffic situation. The employed motion planner follows a widely accepted methodology that first samples candidate trajectories in the trajectory space, then select the one with minimal cost as the planned trajectory. Moreover, in addition to traditional factors such as comfort, efficiency and safety, the cost function is proposed to incorporate incentive of behavior decision like a human driver, so that both lane change decision and motion planning are coupled into one framework. Two types of lane incentive cost --- heuristic and learning based --- are proposed and implemented. To verify the validity of the proposed method, a data set is developed by using the naturalistic trajectory data of human drivers collected on the motorways in Beijing, containing samples of lane changes to the left and right lanes, and car followings. Experiments are conducted with respect to both lane change decision and motion planning, and promising results are achieved.
\end{abstract}


\thispagestyle{fancy}
\fancyhead{}
\lhead{}
\lfoot{\small\copyright 2020 IEEE. Personal use of this material is permitted. Permission from IEEE must be obtained for all other uses.}
\cfoot{}
\rfoot{}


\section{Introduction}

\IEEEPARstart{L}{ast} decade has witnessed tremendous progress in the field of autonomous driving. Many prototyping vehicles have been demonstrated \cite{2}\cite{39}\cite{40}, Advanced Driving Assistant Systems (ADAS) are being evolved from information and warning towards automated driving \cite{41}, and commercial cars with conditional autonomous functions have been produced. These systems have been developed by following mainly the architectures of mobile robotics. While different from a robot, an autonomous driving system needs not only to fulfill its mission (e.g., arriving at a destination) efficiently and safely, but also follow human driver's behavior, so as to let passengers inside the vehicle and other traffic participants on the road feel comfort and relieved.

This issue can be addressed in the motion planning module, where a desired trajectory is selected from a set of samples. However finding a proper cost function to evaluate candidate trajectories is highly non-trivial, which usually requires a significant amount of hand-engineering by experts, and it is even hard to incorporate the evaluation of likelihood to human driver's behavior to the cost function, which remains an open question. On the other hand, in order to reduce workload of planning for long-term and complex driving tasks, hierarchical architecture has been widely used, thus that behavioral decision and motion sequences are planned at different layers. However this may different with the way of human drivers. Modeling driver's decision making and operation process have been studied extensively in the fields of transportation \cite{19}\cite{20}\cite{24} and ergonomics \cite{25}. In these models, a driver's lane change decision is made after confirmation of feasibility of the maneuver \cite{22}\cite{23}, which means that its behavioral decision and motion planning is not separated.

Inspired by the recent efforts on naturalistic driving data collection \cite{42} and analysis \cite{44}\cite{45}\cite{46} and the promising progress on learning from demonstration \cite{49}\cite{50}\cite{Driggs2018HG}, this research proposes a motion planning method by learning from naturalistic data aiming at human-like autonomous driving on crowded highway scenes. At a certain driving condition, a set of trajectory samples is first generated containing both longitudinal and lateral movements, a desired one is then selected with a cost function, which is designed with not only the factors on such as the efficiency, comfort etc., as an existing autonomous driving system does \cite{14}\cite{16}, but also concerns incentive of the decision like a human driver \cite{28}, so as to couple both lane change decision and motion planning into one framework. A method is developed of learning parameters of the cost function by minimizing the spatial-temporal distance between the planned and the human driving trajectories at the same driving situation, which avoids the multiplicity problem of the methods by matching feature expectations \cite{6}\cite{7}\cite{8}. In the authors' early works \cite{47}\cite{48}, a data set is developed through on-road naturalistic driving by human drivers, which contains large sets of longitudinal and lateral driving samples with both the ego and environment vehicle trajectories. Experiments are conducted on these data, and performance of the proposed method is demonstrated.

This paper is structured as follows. A review to the literature works on driving behavior modeling, motion planning and learning from data is given in section II. The outline of the proposed algorithm is described in section III. Different types of trajectory costs including the proposed lane incentive cost are described in section IV. Some details of the algorithm are introduced in section V. Experimental results using on-road driving data is presented in section VI, followed by conclusion and future works in section VII.

\section{Literature Review}

\subsection{Driving Behavior Modeling}

Modeling driving behaviors have gathered significant attention throughout the past decades in the fields of transportation \cite{19}\cite{20}\cite{24} and ergonomics \cite{25} studies. Many of the early works were primarily motivated by the needs of improving road safety and the transportation system as a whole. Recent works have been more focused on looking into details of each driver, vehicle and environment, and incorporate the models into microscopic traffic simulations or Advanced Driver Assistance Systems (ADAS). A commonsense underlying these efforts is that understanding a human's decision making and operation process within a local driving environment of the complex transportation system is of paramount importance. 

Normally driving tasks on a multi-lane motorway are conducted by two fundamental maneuvers: lane keeping and lane changing. The former concerns mainly on a driver's speed or acceleration control, where car following has been extensively studied \cite{21} of the behaviors when the ego's movement is constrained by a front vehicle. The later has been studied mainly on drivers' lane changing decision \cite{22}\cite{23}, which is concerned as the result of three factors: whether it is necessary, desirable and feasible to change lanes \cite{26}. According to its reason, mandatory (MLC) and discretionary (DLC) lane changings are classified. MLC is performed when the driver must leave the current lane, whereas DLC is to improve driving conditions \cite{29}. Many lane changing decision models have been developed in literature, e.g., it is described in a three-stage process: whether or not to make a lane change, target lane choice, and gap acceptance check for executing the lane changing \cite{27}; modeled on incentive and safety criteria: the target lane is more attractive if the incentive criterion is met, the lane change is approved if the safety criterion is met \cite{28}; extended to model the trade-off between the decisions of lane changing and car following \cite{30}\cite{31}. A major limitation of these models is that they fail to capture drivers' path planning and anticipation capabilities over time. Although \cite{32}\cite{33} incorporated prediction of future actions in traffic flow simulation, the kinematic and dynamic constraints of real vehicles were ignored.

\subsection{Motion Planning}

The past decade has witnessed tremendous progress in the field of autonomous driving. Many fully autonomous vehicles are themselves cognitive systems that have their own perception, planning and control modules. In order to reduce the workload of planning for long-term and complex driving tasks, hierarchical architecture has been widely used \cite{11}\cite{2}\cite{17}\cite{14}. For example, \cite{11} decompose planning into hierarchical layers dealing with the tasks at mission, behavior and motion levels, where mission planning returns an optimal route to the destination, behavior executive makes tactical decisions on such as car following or lane changing, motion planning generates a desired trajectory concerning the vehicle's kinematic and dynamic constraints, and output for control module's execution. 

Such hierarchical architecture has a shortcoming on that the higher-level decision making module usually does not have enough detailed information, and the lower-level layer does not have authority to reevaluate the decision \cite{14}. For example, the behavior executive make a decision of lane changing, whereas the motion planner may fail to find a feasible trajectory to fulfill the lane changing mission. To solve this problem, \cite{14} integrated behavioral decision and motion planning into one layer by using a prediction engine. After sampling candidate strategies that contain both longitudinal and lateral movements, the prediction engine forward-simulates each candidate to get trajectories of the subject as well as environmental vehicles. These candidates are finally evaluated by a cost function, and the best one is forwarded to the corresponding ACC or lateral controllers for execution. 

Motion planning of a robotic system is generally framed as finding the lowest cost one from all trajectory candidates \cite{12}\cite{13}\cite{16}\cite{36}. A cost function encodes a system's preference, therefore has significant impact on its performance. However designing a proper cost function could be highly non-trivial, which is usually hand crafted \cite{14}\cite{16}, needs to balance the contributions (e.g., weights) of many components that could potentially be correlated or even contradictory, and it is even harder to design a correct setting to generalize enough at various scenarios \cite{1}\cite{3}. In different with many other robotics, an autonomous driving system needs not only to fulfill its mission (e.g., achieving a destination efficiently and safely), but also follow human driver's behavior, so as to let the passengers inside the autonomous vehicle or other traffic participants of the road feel comfort and relieved. Designing a cost function that encodes human driver's preference in decision making and operation is of great importance \cite{34}\cite{35}\cite{37}\cite{38}.

\subsection{Learning from Data}

Machine learning methods have been used to learn cost functions or parameter settings from the data of human demonstration, where an extensive review to the works on general robotics can be found in \cite{10}. Research efforts are also addressed for autonomous driving applications. The Maximum Margin Planning (MMP) framework is exploited to learn a cost function addressing the coupled problem of both terrain and driving preferences \cite{1}. The problem of learning driving style from expert demonstration is formulated using inverse reinforcement learning (IRL) \cite{6}, and adapted to learn a cost function of path planning for an application of parking lot navigation \cite{3}. The method is then extended to maximum entropy formulation \cite{7}, and exploited to compute trajectories that mimic the driving styles of demonstrators for autonomous driving on highways \cite{8}. In these methods, a cost function is mapping from a linear combination of the global features of a trajectory to costs, and the goal of learning is to find the weights by matching feature expectations between the best trajectory on recovered reward function and the empirical data by human demonstrations. 

However, a problem of these methods is that matching of feature expectation does not guarantee matching of trajectories, i.e., the best trajectory selected may largely different with the one of human demonstration at the same situation, and at complex scenarios like driving at crowded traffics, feature expectations can not be obtained reliably by simply taking averages over a limited number of observations that is used in prior works. Recently, deep learning method has been developed. \cite{9} learnt cost functions by mapping raw sensor measurements to actions using a Fully Convolutional Neural Network that represents more rich features. However the method did not account for the dynamic constraints of robots, and the features on such as speed, curvatures of the trajectories are neglected. Such model-free methods seek to learn a mapping from states to actions, without assuming any prior knowledge of both the system and environment. However, they come at the cost of difficulty with both generalization to new problems, and sequentially combining decisions to achieve longer horizon planning \cite{10}.

There are a type of methods towards human-like planning that use machine learning or data-driven technique in subcomponents. \cite{Driggs2017Integrating} developed a data-driven set prediction method adopted on surrounding vehicles to estimate a inhabit region for future lane changing trajectory, which enables the planner to generate human-like cooperative lane changing behavior. \cite{Driggs2016Communicating} made statistical analysis to get expected trajectory pattern of human driving vehicles before executing lane change, so that planner is encouraged to follow the pattern to mimic human driver's ``preparing to lane change'' behavior. Although these methods also learn from data, they cannot, and actually are not designed, to solve the problem of hand-engineering of trajectory cost, because only a component of the cost are learned in these methods rather than parameters for constituting the overall cost.

\begin{figure*}[tb]
 \begin{center}
    \includegraphics[keepaspectratio=true,width=1\linewidth]{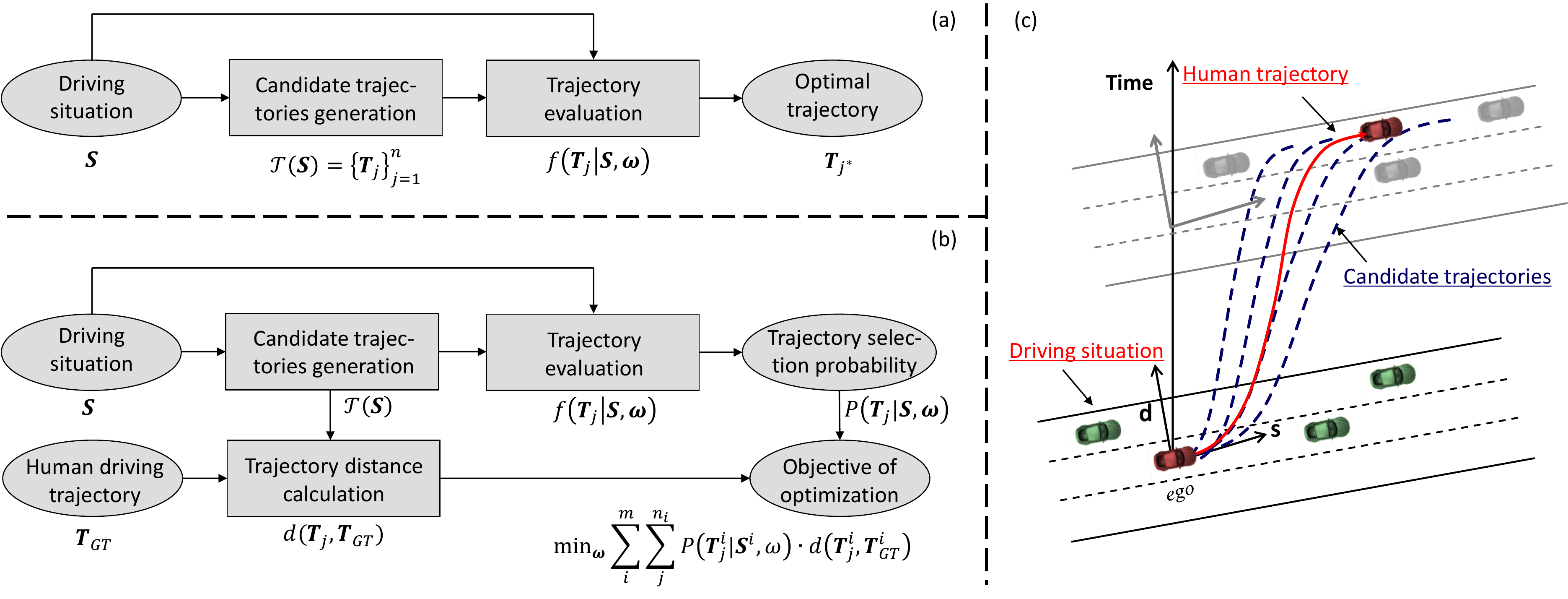}
  \end{center}
\caption{(a) Traditional trajectory planning method. (b) Proposed method of learning human-like trajectory planning. (c) A human driving sample.}
\label{Proposed}
\end{figure*}

\section{Outline of the Algorithm}
\label{problem_formulation}

At a certain driving situation $\boldsymbol{S}$, a traditional motion planning method will first generate a set of synthesized trajectories ${\cal T}(\boldsymbol{S})=\{{\boldsymbol{T}}_1,{\boldsymbol{T}}_2,\ldots,{\boldsymbol{T}}_n\}$, and then select an optimal one for the autonomous driving system's execution, i.e., 
\begin{equation} \label{planning_formula}
\boldsymbol{T}_{j^*}, j^*=\textrm{argmin}_j f(\boldsymbol{T}_j|\boldsymbol{S},\boldsymbol{\omega})
\end{equation} 
where $f$ is a cost function with $\boldsymbol{\omega}$ as its parameters which are usually a set of weights that balance the importance of various considered factors. However, the parameters are usually manually adjusted so that the selected trajectory may not follow human driver's behavior. As illustrated in Fig.~\ref{Proposed}, this research exploits the same work flow, but learns a cost function $f$ that has preference to the trajectories ${T}_j$ that are similar with the human driving ones.

In this work, a human driving sample with index $i$ is denoted as $\boldsymbol{H}^i = (\boldsymbol{S}^i, \boldsymbol{T}_{GT}^i)$, where $\boldsymbol{S}^i$ represents the driving situation at the starting moment of planning, and $\boldsymbol{T}_{GT}^i$ represents the ground truth human driving trajectory in the planning horizon.
Let ${\cal T}(\boldsymbol{S}^i)=\{{\boldsymbol{T}}_1^i,{\boldsymbol{T}}_2^i,...,{\boldsymbol{T}}_{n_i}^i\}$ be the set of candidate trajectories that are generated at driving situation $\boldsymbol{S}^i$ by a certain trajectory generator used in traditional motion planning methods, and let $d(\,\cdot\,,\,\cdot\,)$ represent a measure evaluating distance between trajectories. 

We expect that the trajectory planner is able generate human-like behavior based on cost function $f$, which require us to optimize $\boldsymbol{\omega}$ to make the selected trajectories based on $f$ are as close to the human-driving ones as possible. However, taking this target straightforwardly as the optimization goal will lead to a piece-wise constant objective function which is difficult to optimize. To this end, we propose to modify the selection of the optimal trajectory from the deterministic manner into a probabilistic manner, i.e., $f(\boldsymbol{T}_j|\boldsymbol{S},\boldsymbol{\omega})$, the cost of trajectory $\boldsymbol{T}_j$ at driving situation $\boldsymbol{S}$, is first mapped to the probability of selecting $\boldsymbol{T}_j$ from ${\cal T}(\boldsymbol{S})$, and then the target can be naturally stated as: the expectation of distance between the probabilistically selected trajectory and the human driving one should be as small as possible. Denoting the probability of selecting $\boldsymbol{T}_j$ as $P(\boldsymbol{T}_j|\boldsymbol{S},\boldsymbol{\omega})$, the target can be formulated as the following optimization problem:
\begin{equation}
\label{opt_problem}
\min_{\boldsymbol{\omega}} \sum _i^m \sum _j^{n_i} P(\boldsymbol{T}_j^i|\boldsymbol{S}^i,\boldsymbol{\omega})\cdot d({\boldsymbol{T}}_j^i,{\boldsymbol{T}}_{GT}^i)
\end{equation}
In this paper, a softmax transformation is adopted for mapping a cost to a probability:
\begin{equation}
P(\boldsymbol{T}_j|\boldsymbol{S},\boldsymbol{\omega}) =  \frac{e^{-f(\boldsymbol{T}_j|\boldsymbol{S},\boldsymbol{\omega})}}{\sum_k^n e^{-f(\boldsymbol{T}_k|\boldsymbol{S},\boldsymbol{\omega})}}
\end{equation}
which is reasonable by ensuring that trajectory of higher cost will be assigned lower probability of selection. Hence, if $f$ is properly defined to make it differentiable with respect to $\boldsymbol{\omega}$, then the objective function of the optimization problem will also be differentiable with respect to $\boldsymbol{\omega}$, which enables direct use of well developed gradient-based numerical optimization algorithms to solve the problem.

Below, we first introduce the definition of various components of cost including traditional cost and the proposed lane incentive cost, and then present details of human driving sample $\boldsymbol{H}^i$, trajectory set generation ${\cal T}(\boldsymbol{S})$, distance measure $d$, method of cost learning and process of online planning with decision.

\section{Trajectory Cost Formulation}

\subsection{Traditional Costs}
\label{conventional_costs}

Conventionally, costs regarding comfort, efficiency and safety are considered (e.g., \cite{Xu2011A}). Although the detailed definitions of each cost in different literatures are not exactly the same, they generally follow similar principles. Below is the definition of traditional costs implemented in this study, where a generated trajectory $\boldsymbol{T}_j$ is assumed to start at time $0$ with duration $\tau_j$. The notation will also be applied in \ref{lane_incentive_cost}.

\subsubsection{Comfort}

High acceleration and high acceleration change ratio (i.e., jerk) could be the major reasons of uncomfort.
In Fren\'{e}t frame, vehicle motion is decomposed into longitudinal and lateral movements.
Comfort of a trajectory is evaluated on its acceleration and jerk on two individual dimensions as formulated below.
\begin{eqnarray}
			 c_{lon,jerk}(\boldsymbol{T}_j) &=& \frac{1}{\tau_j}\int_{0}^{\tau_i}|\dddot{s}_j(t)|\,dt \\
			 c_{lat,jerk}(\boldsymbol{T}_j) &=& \frac{1}{\tau_j}\int_{0}^{\tau_i}|\dddot{d}_j(t)|\,dt \\
			 c_{lon,acc}(\boldsymbol{T}_j) &=& \frac{1}{\tau_j}\int_{0}^{\tau_i}|\ddot{s}_j(t)|\,dt \\
			 c_{lat,acc}(\boldsymbol{T}_j) &=& \frac{1}{\tau_j}\int_{0}^{\tau_i}|\ddot{d}_j(t)|\,dt
\end{eqnarray}

\subsubsection{Efficiency}

The higher speed, the higher efficiency. Let $\bar{v}_j$ be the average speed of the trajectory, which is estimated in this research by $\bar{v}_j = (s_j(\tau_j) - s_j(0))/\tau_j$, efficiency is formulated as below:
\begin{equation}
c_{v}(\boldsymbol{T}_j|\boldsymbol{S}) =  \dot{s}_{ego}(0) - \bar{v}_j
\end{equation}

\subsubsection{Safety}

Distances between the ego and environmental vehicles at each moment during the course of the trajectory are examined to evaluate its safety.
Motion trajectories of environmental vehicles are predicted using a linear motion model based on the initial states that are described in $\boldsymbol{S}$, and
are represented as follows:
\begin{equation}
\boldsymbol{T}_{env,q}  = \{(s_q^{env}(t),d_q^{env}(t))\}
\end{equation}
where the subscript ``$q$'' denotes the ID of an environmental vehicle, the preceding and back vehicles on the left, right and current lanes are only concerned.
As longitudinal and lateral distances may contribute to different safety costs, they are estimated individually,
$ D_{s,j}^q(t)=(s_j(t) - s_{q}^{env}(t))^2$ and
$ D_{d,j}^q(t)=(d_j(t) - d_{q}^{env}(t))^2$ , and weighed with a parameter $\lambda _{s}$.
An average is taken on these distances to formulate safety cost as below.
\begin{equation}
c_{safe}(\boldsymbol{T}_j|\boldsymbol{S}) = \sum_{q=1}^{Q_{S}} \frac{1}{\tau_j}\int_{0}^{\tau_j}e^{-({\lambda _s D_{s,j}^q(t)} +
 D_{d,j}^q(t))}
\end{equation}

\subsection{Lane Incentive Cost}
\label{lane_incentive_cost}
``Lane incentive'' is used to describe a driver's preference to the a certain lane which is believed to provide a better driving situation. It is usually implemented using a hard decision rule in conventional driving behavior models \cite{28}. In this section, we propose two methods to represent it as a cost term so that it can be integrated into a cost based planning framework seamlessly. Our interested scope doesn't cover cases that the vehicle must leave the current lane for its destination, which is called mandatory lane change \cite{27} and addressed at the strategic level by path planning. Instead, the discretionary lane change is considered, where a different lane is selected when the current driving condition is not satisfiable (e.g., due to a slow preceding vehicle) and the new target lane has the potential of improving driving condition. 

\begin{figure}[tb]
  \begin{center}
    \includegraphics[keepaspectratio=true,width=0.85\linewidth]{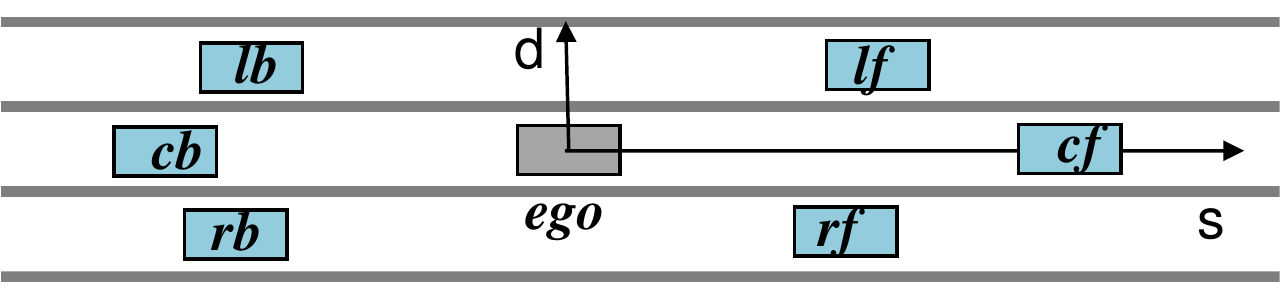}
  \end{center}
\caption{Relevant scene vehicles.}
\label{rel_veh}
\end{figure}

Fig.~\ref{rel_veh} shows relevant scene vehicles that should at least be considered in a discretionary lane change, i.e., ego's closest preceding and rear vehicles on ego's lane and neighbouring lanes. Notations refer to different scene vehicles are also shown in the figure. For example, ``cf'' represent the ``front'' car in the ``center'' lane. Since the relative motion states of relevant scene vehicles are commonly believed to influence a driver's lane change motivation and used to build up a lane change model, we will also use this information to define the lane incentive cost. A notation for relative motion states is introduced as below, which will be used later:
\begin{equation}
\label{s_rel}
s_*^{rel}(t) = s_*(t) - s_{ego}(t)
\end{equation}

Two types of lane incentive cost is introduced in order to model the process of lane change decision. The first type is based on traditional lane change models, which explicitly and heuristically incorporate factors that may motivate lane change decision. The other type is based on learning based behavior model, which implicitly involves the factors by taking all potentially relevant information as input. 

\subsubsection{Heuristic cost} \label{heuristic_lane_incentive_cost}
Conventionally, incentive of lane change is caused by driver's dissatisfaction with the current driving situation, i.e., the preceding vehicle drives slower than expected. Meanwhile, the driver will be attracted by a neighbouring lane if the driving situation there is better, i.e., the ego vehicle is more likely to travel faster if changing to that lane. 

Based on the above understanding, a group of lane incentive costs based on longitudinal relative velocity is defined as follows, which considers driving situations at both start and end point of planned trajectory and rear vehicle is also incorporated to allow for the human driver's polite lane change behavior:
\begin{eqnarray}
             c_{s,tar,f}(\boldsymbol{T}_j|\boldsymbol{S}) &=& -\dot{s}^{rel}_{tar,f}(0)  \\
             c_{s,tar,b}(\boldsymbol{T}_j|\boldsymbol{S}) &=& \dot{s}^{rel}_{tar,b}(0) \\
			 c_{e,tar,f}(\boldsymbol{T}_j|\boldsymbol{S}) &=& -\dot{s}^{rel}_{tar,f}(\tau_j) \\ 
             c_{e,tar,b}(\boldsymbol{T}_j|\boldsymbol{S}) &=& \dot{s}^{rel}_{tar,b}(\tau_j)
\end{eqnarray}
In the above definition, each potential target lane is evaluated individually, including the left, right and current lanes, and denoted by a subscript ``$tar$''.
Velocities of the preceding and back vehicles at the initial time are given in the driving situation $\boldsymbol{S}$, while those at the terminal time are predicted by a linear motion model. 

\subsubsection{Learning based cost}
\label{machine_learning_based_cost}

Instead of explicitly using the knowledge to design a cost, we propose to define the lane incentive cost based on the output of a learning based lane change model, where subtle clues that cannot be observed or formulated manually may be discovered and modeled. We first represent the lane incentive using a lane selection probability output by a learned model, and then map the probability to a lane incentive cost. Each trajectory will be assigned a lane incentive cost according to the target lane it belongs to. Note that in this way, different trajectories can be assigned exactly the same lane incentive cost if their end points fall in the same lane. 

First, the driving situation is encoded into a descriptor of fixed length serving as the input of a learning based classifier. The driving situation descriptor is constructed by concatenating ego's longitudinal velocity and relative motion states of relevant vehicles at planning time (refer to Fig.~\ref{rel_veh} and Eqn.~(\ref{s_rel})):
\begin{equation}
\begin{split}
\boldsymbol{E}(\boldsymbol{S}) = (&\dot{s}_{ego}, |s^{rel}_{cf}|, \dot{s}^{rel}_{cf}, |s^{rel}_{cb}|, \dot{s}^{rel}_{cb}, |s^{rel}_{lf}|, \dot{s}^{rel}_{lf}, \\
        &|s^{rel}_{lb}|, \dot{s}^{rel}_{lb}, |s^{rel}_{rf}|, \dot{s}^{rel}_{rf}, |s^{rel}_{rb}|, \dot{s}^{rel}_{rb})
\end{split}
\end{equation}

In case that some of the scene vehicles don't exist, corresponding virtual vehicles will be introduced, whose longitudinal distance is set to a large default value and the longitudinal relative velocity is set to a value with large default absolute value whose sign is determined according to whether the virtual vehicle is a leading one (positive sign) or a rear one (negative sign). This operation is reasonable because from a physical point of view, the introduced virtual vehicle is so far and getting further from the ego vehicle that it cannot influence the decision of the ego vehicle. Meanwhile, from a algorithmic point of view, the motion state of a virtual vehicle is quite different from that of a real existing vehicle, so that it's simple for a classifier to distinguish the situation where a virtual vehicle is introduced. 

In case that a left of right lane doesn't exist, a virtual lane is introduced with two virtual vehicles whose longitudinal distance and longitudinal relative velocity are both set to $0$. This operation is reasonable because from a physical point of view, the virtual situation is actually a extreme case that lane change cannot be executed because that gap between the virtual leading vehicle and the rear one is $0$. Meanwhile, from a algorithmic point of view, the virtual case will never happen if a neighbouring lane exists, so that it's simple for a classifier to distinguish the situation where a left or right neighbouring lane doesn't exist.

\begin{figure}[tb]
  \begin{center}
    \includegraphics[keepaspectratio=true,width=1\linewidth]{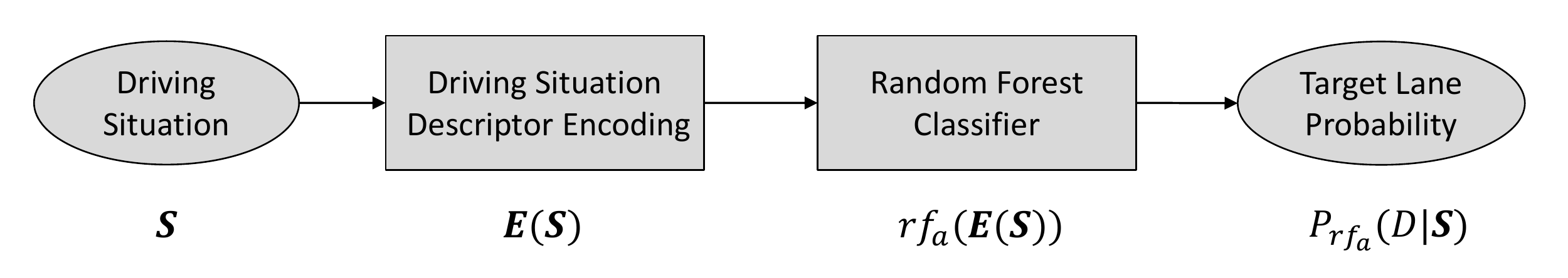}
  \end{center}
\caption{Pipeline of evaluating the probability of lanes selected as the target based on a random forest.}
\label{rf}
\end{figure}

Then, a classifier is trained to map scene descriptors to human-like decisions for lane selection (denoted as $D$). We propose two ways of defining the decision pool: 1) two-way decisions, i.e., lane change (LC) and car following (CF); 2) three-way decisions, i.e., left lane change (LLC), right lane change (RLC) and car following. A random forest model \cite{RF} is adopted as the classifier, which is proved to be effective in driver intent identification \cite{Driggs2015Identifying}, and its hyper-parameter, minimal number of samples in leaf nodes, is determined using a cross-validation technique on the training set. Two random forest models are trained based on two decision pools respectively, and the random forest trained based on the two-way decisions is denoted as $rf_2$ while the model corresponding to three-way decisions is denoted as $rf_3$, i.e.,
\begin{equation}
\begin{split}
rf_2:&\ D\in\{\text{LC, } \text{CF}\} \\
rf_3:&\ D\in\{\text{LLC, } \text{CF, }\text{RLC}\}
\end{split}
\end{equation} 
Thus, the probability of generating a lane decision $D$ in driving situation $\boldsymbol{S}$ can be estimated by the random forest $rf_a, a=2,3$, which is denoted as $P_{rf_a}(D|\boldsymbol{S})$. Fig.~\ref{rf} shows the pipeline of generating such a probability. Based on the estimated probability, the lane incentive cost of trajectory $\boldsymbol{T}_j$ in driving situation $S$ is defined as:
\begin{equation}
c_{rf_a}(\boldsymbol{T}_j|\boldsymbol{S})=-\log(P_{rf_a}(D_j|\boldsymbol{S}))
\end{equation}
where $D_j$ is the lane decision of $\boldsymbol{T}_j$, which is determined by position of the end point of $\boldsymbol{T}_j$.

\subsection{Total Cost}
\label{subsec_total_cost}
In this study, four versions of total cost $f_0(\,\cdot\,|\,\cdot\,,\boldsymbol{\omega}_0)$, $f_1(\,\cdot\,|\,\cdot\,,\boldsymbol{\omega}_1)$, $f_2(\,\cdot\,|\,\cdot\,,\boldsymbol{\omega}_2)$ and $f_3(\,\cdot\,|\,\cdot\,,\boldsymbol{\omega}_3)$ are defined, which will be referenced in experiments. They are different in the way that whether the total cost contains the lane incentive cost and if so, which version of lane incentive cost is involved. Our definition of total cost basically follows the common approach, i.e., weighted summation of every single cost term. In this way, the weighting parameters are what need to be learned from data, i.e., $\boldsymbol{\omega}$ in $f$ (refer to \ref{problem_formulation}). However, there is an extension introduced in our definition: powers of each cost term are introduced as new cost terms of which each will be assigned an independent weighting parameter. This variation gives the cost function more degrees of freedom because extra independent weighting parameters are introduced for powers of original cost terms, so that the cost function is potentially able to fit the latent human driver cost better and produce human-like behaviors.

Let $\boldsymbol{C}_{trad}(\boldsymbol{T}_j|\boldsymbol{S})$ be a column vector formed by concatenating all terms of traditional trajectory costs (refer to \ref{conventional_costs}), i.e.,
\begin{equation}
\boldsymbol{C}_{trad} = (c_{lon,jerk},c_{lat,jerk},c_{lon,acc},c_{lat,acc},c_v,c_{safe})^T
\end{equation}
$\boldsymbol{C}_{trad}(\boldsymbol{T}_j|\boldsymbol{S})^{(k)}$ be the vector of extended cost terms which are $k$ power of original cost terms, i.e., 
\begin{equation}
\boldsymbol{C}_{trad}^{(k)} = (c_{lon,jerk}^k,c_{lat,jerk}^k,c_{lon,acc}^k,c_{lat,acc}^k,c_v^k,c_{safe}^k)^T
\end{equation}
and $\boldsymbol{\omega}_{trad}^{(k)}$ be a row vector of the same length, containing independent weighting parameters of cost terms in $\boldsymbol{C}_{trad}(\boldsymbol{T}_j|\boldsymbol{S})^{(k)}$. With these notations, the total cost with only traditional terms is defined as follows:
\begin{equation} \label{total_cost_0}
f_0(\boldsymbol{T}_j|\boldsymbol{S},\boldsymbol{\omega}_0) = \sum_{k=1}^K \boldsymbol{\omega}_{trad}^{(k)} \boldsymbol{C}_{trad}(\boldsymbol{T}_j|\boldsymbol{S})^{(k)}
\end{equation} 
where $\boldsymbol{\omega}_0$ represents the collection of all weighting parameters, i.e., 
\begin{equation}
\boldsymbol{\omega}_0=(\boldsymbol{\omega}_{trad}^{(1)},\boldsymbol{\omega}_{trad}^{(2)},\cdots,\boldsymbol{\omega}_{trad}^{(K)})
\end{equation}

The second version of total cost is composed of traditional cost terms and heuristic lane incentive cost terms (refer to \ref{heuristic_lane_incentive_cost}). Let $\boldsymbol{C}_{heu}(\boldsymbol{T}_j|\boldsymbol{S})$ be the column vector formed by concatenating all terms of heuristic lane incentive costs, i.e.,
\begin{equation}
\boldsymbol{C}_{heu} = (c_{s,tar,f},c_{s,tar,b},c_{e,tar,f},c_{e,tar,b})^T
\end{equation}
Similar to notations in Eqn.~(\ref{total_cost_0}), $\boldsymbol{C}_{heu}(\boldsymbol{T}_j|\boldsymbol{S})^{(k)}$ and $\boldsymbol{\omega}_{heu}^{(k)}$ can be defined, and the total cost of this version is defined as follows:
\begin{equation} \label{total_cost_1}
\begin{split}
f_1(\boldsymbol{T}_j|\boldsymbol{S},\boldsymbol{\omega}_1) &= \sum_{k=1}^K \boldsymbol{\omega}_{trad}^{(k)} \boldsymbol{C}_{trad}(\boldsymbol{T}_j|\boldsymbol{S})^{(k)} \\
&+ \sum_{k=1}^K \boldsymbol{\omega}_{heu}^{(k)} \boldsymbol{C}_{heu}(\boldsymbol{T}_j|\boldsymbol{S})^{(k)}
\end{split}
\end{equation} 
where
\begin{equation}
\boldsymbol{\omega}_1=(\boldsymbol{\omega}_{trad}^{(1)},\cdots,\boldsymbol{\omega}_{trad}^{(K)},\boldsymbol{\omega}_{heu}^{(1)},\cdots,\boldsymbol{\omega}_{heu}^{(K)})
\end{equation}

The third and fourth version of total cost are defined by replacing the heuristic lane incentive cost terms in the second version with the random forest based lane incentive cost term (refer to \ref{machine_learning_based_cost}) with two-way and three-way decisions respectively. Note that there is only one related cost term, so we don't need to define the cost vector in this case. Let $\omega_{rf}$ be the weighting parameter for this cost term, the total cost of these two versions is defined as:
\begin{equation}
\begin{split}
f_2(\boldsymbol{T}_j|\boldsymbol{S},\boldsymbol{\omega}_2) &= \sum_{k=1}^K \boldsymbol{\omega}_{trad}^{(k)} \boldsymbol{C}_{trad}(\boldsymbol{T}_j|\boldsymbol{S})^{(k)} \\
&+ \omega_{rf_2}c_{rf_2}(\boldsymbol{T}_j|\boldsymbol{S})
\end{split}
\end{equation}
\begin{equation}
\begin{split}
f_3(\boldsymbol{T}_j|\boldsymbol{S},\boldsymbol{\omega}_3) &= \sum_{k=1}^K \boldsymbol{\omega}_{trad}^{(k)} \boldsymbol{C}_{trad}(\boldsymbol{T}_j|\boldsymbol{S})^{(k)} \\ 
&+ \omega_{rf_3}c_{rf_3}(\boldsymbol{T}_j|\boldsymbol{S})
\end{split}
\end{equation}
where
\begin{eqnarray}
\boldsymbol{\omega}_2=(\boldsymbol{\omega}_{trad}^{(1)},\cdots,\boldsymbol{\omega}_{trad}^{(K)},\omega_{rf_2})\\
\boldsymbol{\omega}_3=(\boldsymbol{\omega}_{trad}^{(1)},\cdots,\boldsymbol{\omega}_{trad}^{(K)},\omega_{rf_3})
\end{eqnarray}


\section{Algorithm Details}

\subsection{Human Driving Samples}

Each data sample $\boldsymbol{H}=(\boldsymbol{S},\boldsymbol{T}_{GT})$ is described in a Fren\'{e}t frame \cite{Werling2010Optimal} with the origin at the ego vehicle's location at the initial time. Let $\boldsymbol{p}=(s,d)$ be a location at the Fren\'{e}t frame, where $s$ and $d$ are displacements from the origin on longitudinal and lateral road directions, and $\dot{\boldsymbol{p}}=(\dot{s},\dot{d})$ and $\ddot{\boldsymbol{p}}=(\ddot{s},\ddot{d})$ are velocity and acceleration vectors.
A driving situation ${\boldsymbol{S}}=\{ {\boldsymbol{S}}_{ego}, {\boldsymbol{S}}_{env}, {S}_{road}\}$ is described with three components, where the first two are the states of the ego and environmental vehicles at the local surrounding that are described by their location, velocity and acceleration as below,
\begin{eqnarray}
&& \boldsymbol{S}_{ego}=(\boldsymbol{p},\dot{\boldsymbol{p}},\ddot{\boldsymbol{p}})_{ego}\\
&& \boldsymbol{S}_{env}=\{(\boldsymbol{p},\dot{\boldsymbol{p}},\ddot{\boldsymbol{p}})_{env}^j|j=1,..,n_{env}\}
\end{eqnarray}
where $j$ is index of the environmental vehicles. In this study, the preceding and back vehicles on the left, right and current lanes are concerned.
${S}_{road} \in \{-1,0,1\}$ is about road situation, indicating that the ego vehicle is driving at the farthest left lane, a middle lane or the farthest right lane.
As for the human driving trajectory ${\boldsymbol{T}}_{GT}$, it is a time series of trajectory points $\{\boldsymbol{P}_k|0 \le k\Delta t \le \tau\}$, where $\boldsymbol{P}_k=(\boldsymbol{p},\dot{\boldsymbol{p}},\ddot{\boldsymbol{p}})_{k}$, $\tau$ is the duration of the trajectory, and $\Delta t$ is the time interval of data sampling. In case of a lane change sample, the duration is determined by the natural start point and end point of the behavior, which usually distributes between 6s-8s as analyzed in \cite{48}. In case of car following samples, a fixed duration ($\tau=8s$ in this study) is adopted.

\subsection{Generation of Candidate Trajectories} \label{subsec_candi_trajs}

\begin{figure}[tb]
  \begin{center}
    \includegraphics[keepaspectratio=true,width=0.85\linewidth]{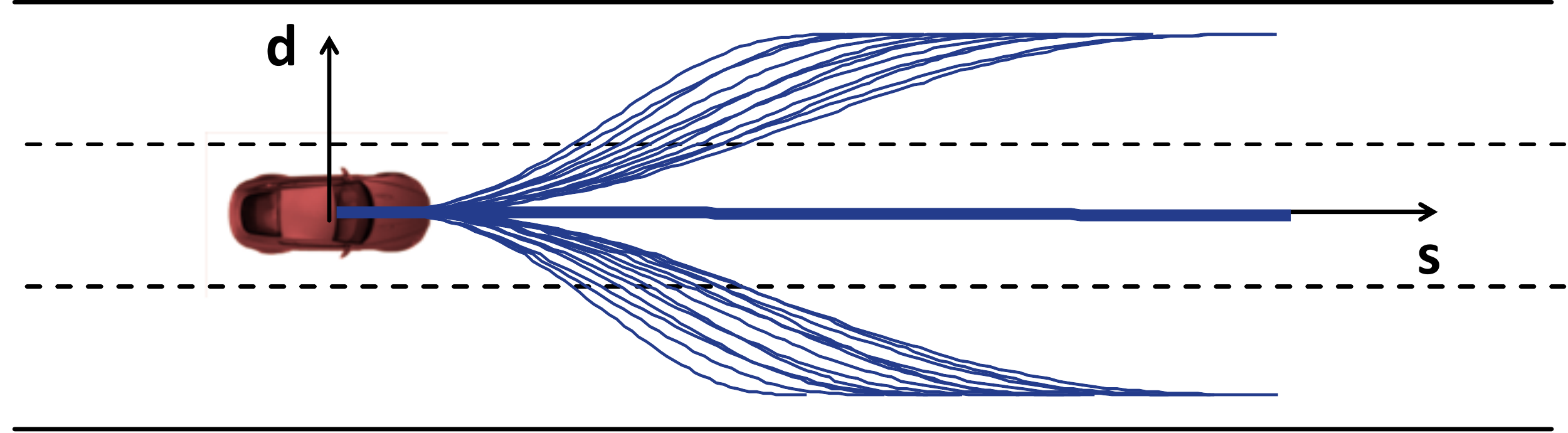}
  \end{center}
\caption{Generated candidate trajectories. The lane keeping trajectories overlap with each other.}
\label{candi_trajs}
\end{figure}

At a certain driving situation $\boldsymbol{S}$, a set of trajectories ${\cal T}(\boldsymbol{S})$ are generated describing potential candidates of future motion sequences. In this research, the trajectory generation method \cite{Xu2011A} is used, which represents a trajectory using two quintic polynomials on lateral and longitudinal axis respectively:
\begin{equation}
\left\{
             \begin{array}{lr}
             d(t)=a_0 + a_{1}t + a_{2}t^2 + a_{3}t^3 + a_{4}t^4 + a_{5}t^5&  \\
             s(t)=b_0 + b_{1}t + b_{2}t^2 + b_{3}t^3 + b_{4}t^4 + b_{5}t^5&
             \end{array}
\right.
\end{equation}

For simplicity, we set $t_s=0$ and subsequently $t_e=\tau$. Given the value of $\tau$ and two trajectory points $\boldsymbol{P} =(\boldsymbol{p},\dot{\boldsymbol{p}},\ddot{\boldsymbol{p}})$ at the initial $t=0$ and terminal times $t=\tau$, six equations can be derived for both $d(t)$ and $s(t)$, consequently the coefficients $\{a_0,...,a_5\}$ and $\{b_0,...,b_5\}$ are estimated.

Assuming that velocity on lateral axis, and acceleration on both longitudinal and lateral axes are zero at both the initial and end times are 0, and since the trajectory point at the initial time is given by ${\boldsymbol{S}}$, which has its location at the origin of the defined Fren\'{e}t frame, we have the following known conditions,
\begin{equation} \label{stedstatecon}
\left\{
             \begin{array}{lr}
             d(0)={\boldsymbol{S}}_{ego}.d,\dot{d}(0)=0,\ddot{d}(0)=0,\dot{d}(\tau)=0,\ddot{d}(\tau)=0\\
             s(0)={\boldsymbol{S}}_{ego}.s,\dot{s}(0)={\boldsymbol{S}}_{ego}.\dot{s},\ddot{s}(0)=\boldsymbol{S}_{ego}.\ddot{s},\ddot{s}(\tau)=0
             \end{array}
\right.
\end{equation}

Highway driving has some properties that can be used in trajectory generation: lane width ($D_{lane}$) and speed limits ($V_{max}$) at a certain road are known; lateral vehicle positions at the initial and terminal time of each maneuver are usually at the middle of lanes; longitudinal displacement could vary largely with respect to different velocity at the initial time, while longitudinal speed may vary only slightly ($\Delta V$) to keep constant and smooth driving. Based on the prior knowledge, terminal states are tessellated within a limited space shown as follows:
\begin{equation}
\left\{
             \begin{array}{lr}
             d(\tau) \in \{-D_{lane},0,D_{lane}\}\\
             \dot{s}(\tau) \in [\max({\boldsymbol{S}}_{ego}.\dot{s} - \Delta V,0),\min({\boldsymbol{S}}_{ego}.\dot{s}+ \Delta V,V_{max})]\\
             \tau \in [\tau_{min},\tau_{max}]
             \end{array}
\right.
\label{sampledetails}
\end{equation}
Fig.~\ref{candi_trajs} shows an example of candidate trajectories generated following the way mentioned above.

\subsection{Distance Measure between Trajectories}

Given a pair of synthesized trajectory $\boldsymbol{T}_1$ and $\boldsymbol{T}_2$, discretization is first conducted to convert continuous trajectories to sequences of synchronized points at an equal interval $\Delta t$.
\begin{equation}
\boldsymbol{T}_i = \{(\boldsymbol{p}_i(t),\dot{\boldsymbol{p}}_i(t))|t =0, \Delta t,..,n_i\Delta t\}
\end{equation}
where $\boldsymbol{p}_i(t)=(s_i(t),d_i(t))$ is trajectory $\boldsymbol{T}_i$'s location in the Fren\'{e}t frame at time $t$.

Distance between each pair of synchronized trajectory points are estimated on both the location ($D(t) = || \boldsymbol{p}_1(t)-\boldsymbol{p}_2(t)||$)  and velocity ($\dot{D}(t) = ||\dot{\boldsymbol{p}}_1(t)-\dot{\boldsymbol{p}}_2(t)||$) components. 
Weighing by a parameter $\lambda _{d}$ and taking an average of these distances during the course, a distance measure between ${\boldsymbol{T}}_1$ and ${\boldsymbol{T}}_2$ is formulated below
\begin{equation}
d({\boldsymbol{T}}_1,{\boldsymbol{T}}_2)  = \frac { \sum_{n=1}^{n_{min}} {(D(n\Delta t) + \lambda _{d} \dot{D}(n\Delta t))}}{n_{min}}
\end{equation}
where $n_{min} = \min(n_1,n_2)$.

\subsection{Cost Learning}
Cost parameters $\boldsymbol{\omega}$ are learned by solving the optimization problem shown as Eqn.~(\ref{opt_problem}). We will first show that gradient of objective function with respect to $\boldsymbol{\omega}$ can be obtained so that the problem can be solved using gradient-based numerical optimization algorithms. Then, we explain how to implement the whole learning procedure efficiently.

Without causing any ambiguity, simplified notations are introduced in the remaining part of the section: $d_j^i$ representing $d(\boldsymbol{T}_j^i, \boldsymbol{T}_{GT}^i)$, $f_j^i(\boldsymbol{\omega})$ for $f(\boldsymbol{T}_j^i|\boldsymbol{S}^i,\boldsymbol{\omega})$ and $P_j^i(\boldsymbol{\omega})$ for $P(\boldsymbol{T}_j^i|\boldsymbol{S}^i,\boldsymbol{\omega})$. Using the notations and according to Eqn.~(\ref{opt_problem}), the objective function to be minimized can be written as:
\begin{equation}
L(\boldsymbol{\omega}) = \sum_i^m\sum_j^{n_i}P_j^i(\boldsymbol{\omega})d_j^i
\end{equation}
To calculate the gradient $\partial L(\boldsymbol{\omega})/\partial \boldsymbol{\omega}$, we need first to get $\partial P_j^i(\boldsymbol{\omega})/\partial \boldsymbol{\omega}$:
\begin{equation}
\begin{split}
\frac{\partial P_j^i(\boldsymbol{\omega})}{\partial \boldsymbol{\omega}}=&\frac{-e^{-f_j^i(\boldsymbol{\omega})}\frac{\partial f_j^i(\boldsymbol{\omega})}{\partial \boldsymbol{\omega}}\sum_k^{n_i}e^{-f_j^i(\boldsymbol{\omega})}}{(\sum_k^{n_i}e^{-f_k^i(\boldsymbol{\omega})})^2} - \\
&\frac{e^{-f_j^i(\boldsymbol{\omega})}\sum_k^{n_i}-e^{-f_k^i(\boldsymbol{\omega})}\frac{\partial f_k^i(\boldsymbol{\omega})}{\partial \boldsymbol{\omega}}}{(\sum_k^{n_i}e^{-f_k^i(\boldsymbol{\omega})})^2}\\
=&\frac{e^{-f_j^i(\boldsymbol{\omega})}\sum_k^{n_i}e^{-f_k^i(\boldsymbol{\omega})}(\frac{\partial f_k^i(\boldsymbol{\omega})}{\partial \boldsymbol{\omega}}-\frac{\partial f_j^i(\boldsymbol{\omega})}{\partial \boldsymbol{\omega}})}{(\sum_k^{n_i}e^{-f_k^i(\boldsymbol{\omega})})^2}\\
=&P_j^i(\boldsymbol{\omega})\sum_k^{n_i}P_k^i(\boldsymbol{\omega})(\frac{\partial f_k^i(\boldsymbol{\omega})}{\partial \boldsymbol{\omega}}-\frac{\partial f_j^i(\boldsymbol{\omega})}{\partial \boldsymbol{\omega}})
\end{split}
\end{equation}
With this result, $\partial L(\boldsymbol{\omega})/\partial \boldsymbol{\omega}$ can be derived as:
\begin{equation}
\begin{split}
\frac{\partial L(\boldsymbol{\omega})}{\partial \boldsymbol{\omega}}=&\sum_i^m\sum_j^{n_i}d_j^i\frac{\partial P_j^i(\boldsymbol{\omega})}{\partial \boldsymbol{\omega}}\\
=&\sum_i^m\sum_j^{n_i}d_j^iP_j^i(\boldsymbol{\omega})\sum_k^{n_i}P_k^i(\boldsymbol{\omega})\left(\frac{\partial f_k^i(\boldsymbol{\omega})}{\partial \boldsymbol{\omega}}-\frac{\partial f_j^i(\boldsymbol{\omega})}{\partial \boldsymbol{\omega}}\right)\\
=&\sum_i^m\sum_j^{n_i}\sum_k^{n_i}d_j^iP_j^i(\boldsymbol{\omega})P_k^i(\boldsymbol{\omega})\left(\frac{\partial f_k^i(\boldsymbol{\omega})}{\partial \boldsymbol{\omega}}-\frac{\partial f_j^i(\boldsymbol{\omega})}{\partial \boldsymbol{\omega}}\right)\\
=&\sum_i^m\sum_j^{n_i}d_j^iP_j^i(\boldsymbol{\omega})\sum_k^{n_i}P_k^i(\boldsymbol{\omega})\frac{\partial f_k^i(\boldsymbol{\omega})}{\partial \boldsymbol{\omega}}-\\
&\sum_i^m\sum_j^{n_i}d_j^iP_j^i(\boldsymbol{\omega})\frac{\partial f_j^i(\boldsymbol{\omega})}{\partial \boldsymbol{\omega}}\sum_k^{n_i}P_k^i(\boldsymbol{\omega})\\
=&\sum_i^m\sum_k^{n_i}d_k^iP_k^i(\boldsymbol{\omega})\sum_j^{n_i}P_j^i(\boldsymbol{\omega})\frac{\partial f_j^i(\boldsymbol{\omega})}{\partial \boldsymbol{\omega}}-\\
&\sum_i^m\sum_j^{n_i}d_j^iP_j^i(\boldsymbol{\omega})\frac{\partial f_j^i(\boldsymbol{\omega})}{\partial \boldsymbol{\omega}}\\
=&\sum_i^m\sum_j^{n_i}P_j^i(\boldsymbol{\omega})\left(\sum_k^{n_i}d_k^iP_k^i(\boldsymbol{\omega})-d_j^i\right)\frac{\partial f_j^i(\boldsymbol{\omega})}{\partial \boldsymbol{\omega}}
\end{split}
\end{equation}
According to total cost definition in \ref{subsec_total_cost}, the total cost in this study is simply weighted sum of all cost terms with $\boldsymbol{\omega}$ as weighting parameters. Thus, $\partial f_j^i/\partial \boldsymbol{\omega}$ is simply the vector concatenating all cost terms of $j$th candidate trajectory in $i$th sample. Denoting this cost vector as $\boldsymbol{C}_j^i$, we write down the final formula for gradient calculation as follows for reference:
\begin{equation}
\frac{\partial L(\boldsymbol{\omega})}{\partial \boldsymbol{\omega}} = \sum_i^m\sum_j^{n_i}P_j^i(\boldsymbol{\omega})\left(\sum_k^{n_i}d_k^iP_k^i(\boldsymbol{\omega})-d_j^i\right)\boldsymbol{C}_j^i
\end{equation}

From this equation, by carefully analyzing each component of right-hand side, it can be found that $d_j^i$ and $\boldsymbol{C}_j^i$ is independent of $\boldsymbol{\omega}$, so that they can be calculated only once and cached for use at any time during optimization. $\sum_k^{n_i}d_k^iP_k^i(\boldsymbol{\omega})$ is independent of $j$, so that when $\boldsymbol{\omega}$ is updated, it can be calculated only once for each sample $i$, and applied to all $j$s. By adopting these measures to reduce repeated computation, the optimization will be much more efficient. In our implementation, L-BFGS algorithm is applied to solve the non-linear optimization problem. 

\subsection{Planning with Decision} \label{subsec_planning_and_decision}
After $\boldsymbol{\omega}$ is learned using the proposed method with collected human-driving data, online human-like planning can be performed according to Eqn.~(\ref{planning_formula}). According to the candidate trajectory generation method described in \ref{subsec_candi_trajs}, end point of each generated trajectory is assumed to fall on one of the center lines of left, ego and right lane. Thus, each planned trajectory corresponds to one of the three behaviors (left lane change, lane keeping and right lane change) determined by its end point's position.  

It's worth noting that the proposed human-like planner can be regarded as a trajectory predictor as well as a lane change maneuver predictor for human driving vehicles. However, it's different from most lane change prediction methods \cite{Lef2014A} in that the proposed model learns to \emph{generate} lane change intention caused by contextual driving situation while most of others essentially \emph{detect} lane change intention mainly depending on observed movement tendency (e.g. \cite{Kumar2013Learning}) when intention has already existed for seconds. In other words, the proposed method is able to predict lane change behavior before the vehicle starts to move towards target lane. Data and experiments in section \ref{sec_exp} actually demonstrate performance of the proposed method for such an prediction application.

\section{Experiments} \label{sec_exp}

\subsection{Dataset} \label{subsec_dataset}
\begin{figure}[tb]
  \begin{center}
    \includegraphics[keepaspectratio=true,width=1\linewidth]{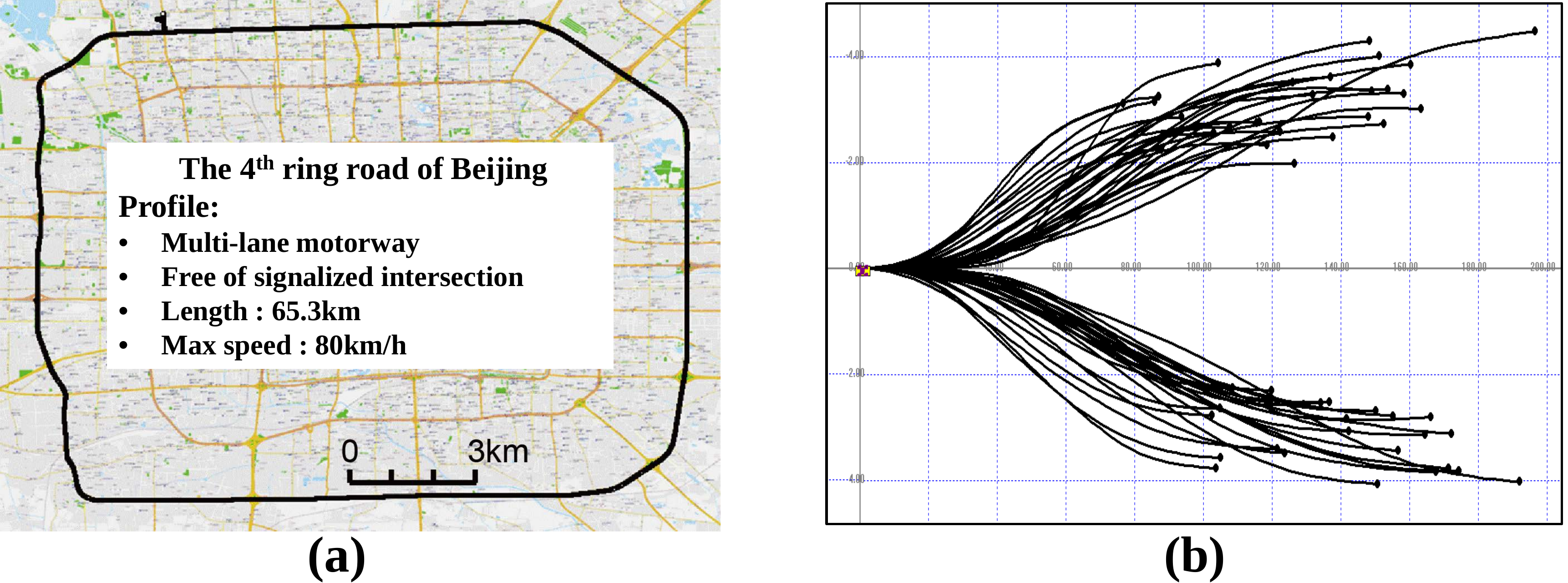}
  \end{center}
\caption{Data Collection Route and Lane Changing Trajectories}
\label{data}
\end{figure}

\begin{figure}[tb]
  \begin{center}
    \includegraphics[keepaspectratio=true,width=1\linewidth]{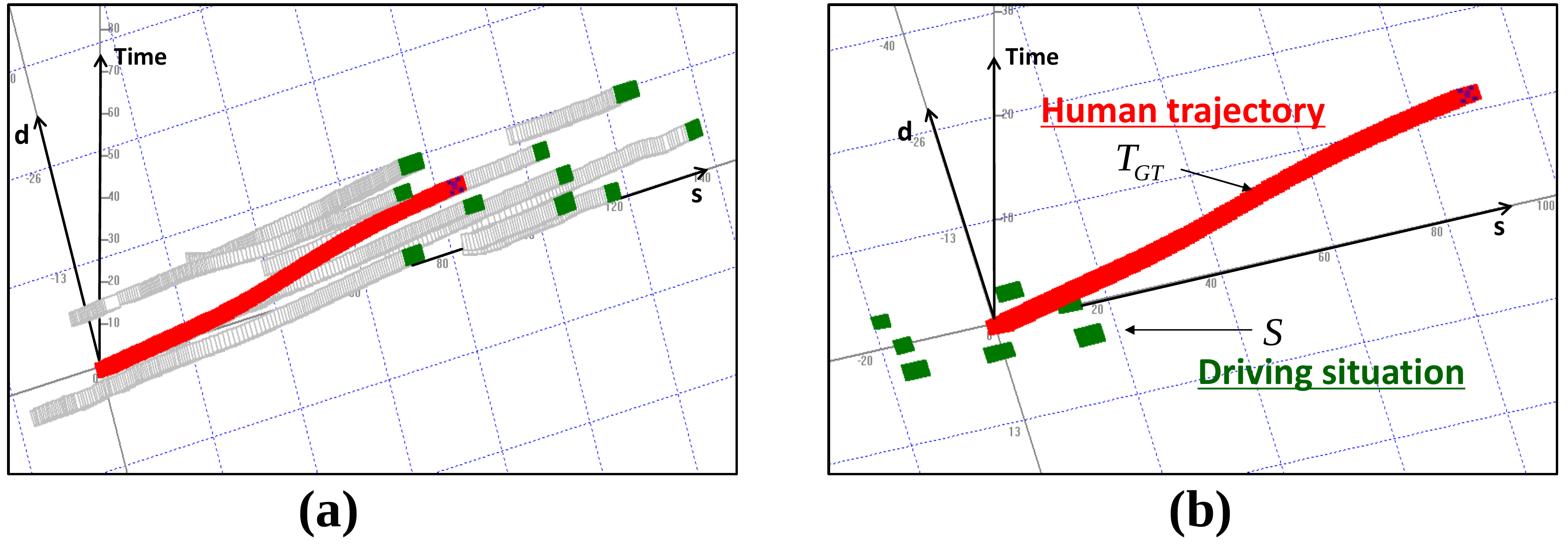}
  \end{center}
\caption{Lane changing Case and Naturalistic human driving Sample}
\label{humanSample}
\end{figure}

\begin{table}[tb]
\begin{center} \caption{DATA PROFILE.}
\label{dataTable}
\renewcommand{\arraystretch}{1.25}
\begin{tabular}{|c|c|c|c|c|}
\hline
Lap & Date & LLC Num& RLC Num & CF Num \\
\hline
1 & 2016/10/12 &9 &13 & 29 \\
\hline
2 &2016/10/15 &8 &11 & 24 \\
\hline
3 &2016/10/17 & 4 & - & 22 \\
\hline
4 &2016/10/19 &- &- & 19\\
\hline
5 &2016/10/22 & 12 & 11 & 15\\
\hline
6 &2016/10/23 &6 &6 & 23\\
\hline
7 &2016/10/24 &3 &2 & 11\\
\hline
8 &2014/01/02 &19 &16 & -\\
\hline
9 &2014/01/02 &7 &7 & -\\
\hline
10 &2014/01/06 &11 &16 & -\\
\hline
11 &2014/01/06 &21 &24 & -\\
\hline
12 &2014/01/06 &18 &16 & -\\
\hline
13 &2014/01/06 &17 &13 & -\\
\hline
\multicolumn{2}{|c|}{Total}&135 &135 & 143\\
\hline
\end{tabular}
\end{center}
\end{table}
A system of on-road vehicle trajectory collection is developed in the authors' early works \cite{47}, which is used in this research to collect on-road naturalistic driving data.
As shown in Fig.~\ref{data}, data collection are conducted on the 4th Ring Road in Beijing. It is a multi-lane motorway that is free of traffic signals, has a total distance of 65.3 km and a designed maximum speed of 80 km/h.
Each data collection means one-round driving (called a ``lap'') on the 4th Ring Road by following traffic flow naturalistically. As listed in Tab.~\ref{dataTable}, human driving samples are extracted from 13 laps of data.
Lane change segments are extracted by using the method of \cite{48}, containing both left and right lane changes.
Car following ones are extracted for a fixed duration of 8s (i.e., an average duration of lane changes) when the following conditions are met simultaneously: 1) driving on straight roads and 2) keeping a distance with its front vehicle within 40m.
Since driving behaviors at jammed conditions could be much different, in this research, the segments with the ego's initial speed lower than 8m/s are discarded.
Fig.\ref{humanSample}(a) shows a lane change segment, which contains both the ego and environment vehicle trajectories during the period of the lane change.
With each segment, the road, ego and environmental vehicles' states at the initial time $t_s$ are used to make $\boldsymbol{S}$ for driving situation, the ego vehicle's trajectory from $t_s$ to $t_e$ is extracted as ${\boldsymbol{T}}_{GT}$ , and the $\boldsymbol{H}=(\boldsymbol{S},\boldsymbol{T}_{GT})$ pair composes a human driving sample as shown in Fig.\ref{humanSample}(b).
From 13 laps of data, a set of human driving samples containing 135 left lane change (LLC), 135 right lane change (RLC) and 143 car following (CF) are developed and used in the experiments below (Tab.~\ref{dataTable}).

\subsection{Experimental Design}

We seek to answer two key questions regarding the objective of the proposed method: 1) whether the learnt planner could propose a trajectory that is close to that of human driving one under the following premises? 2) whether the learnt planner could make a human-like lane change decision under the premises that target lane is undecided?
\begin{itemize}
  \item $Pre.1$: a target lane has been decided;
  \item $Pre.2$: lane change maneuver has been decided, while it could be made to either the left or right lane;
  \item $Pre.3$: subsequently maneuver has not been decided, which could be left or right lane change, or car following.
\end{itemize}

The following experiments are subsequently designed.
\begin{itemize}
  \item $Exp.1$: lane change motion planning to a target lane;
  \item $Exp.2$: lane change motion planning with simultaneous decision of the target lane;
  \item $Exp.3$: motion planning with simultaneous decision of maneuver.
\end{itemize}

In Exp. 1, since there's no process of lane selective decision, there's no way to introduce lane incentive costs so that only $f_0$ is trained and tested. As for Exp. 2, since there's lane change target selection, heuristic lane incentive cost is introduced, i.e., $f_0$ and $f_1$ are implemented, while the learning based lane incentive costs are still excluded because they are trained with the car-following option. Finally in Exp. 3, all the four versions of total costs are implemented and their results are compared and analyzed. The data used in the three experiments are also different because of the experimental setting: in the first two experiments, the human driving samples of RLC and LLC are only used, while all samples are used in the third experiment.
In each experiments, the samples are randomly divided into two groups as listed in Tab.~\ref{expDataTable} for training and testing. Experimental results are presented below.

\begin{table}[tb]
	\centering
	\caption{EXPERIMENT DATA }
	\begin{tabular}{|c|c|c|c|c|c|}
		\hline
        \begin{minipage}{2cm} \vspace{0.1cm}\centering \vspace{0.3cm}\end{minipage}
        {} & Train Num & Test Num           & $Exp.1$& $Exp.2$     &$Exp.3$         \\ \hline
        \begin{minipage}{2cm} \vspace{0.1cm}\centering  \vspace{0.3cm}\end{minipage}
        {CF} & 90 & 53          & $\times$ & $\times$   & $\bigcirc$          \\ \hline
        \begin{minipage}{2cm} \vspace{0.1cm}\centering \vspace{0.3cm}\end{minipage}
		{LLC} & 90 & 45         & $\bigcirc$  & $\bigcirc$   & $\bigcirc$          \\ \hline
        \begin{minipage}{2cm} \vspace{0.1cm}\centering \vspace{0.3cm}\end{minipage}
        {RLC} & 90 & 45         & $\bigcirc$  & $\bigcirc$    & $\bigcirc$    \\ \hline

	\end{tabular}
    \label{expDataTable}
\end{table}

\subsection{Experimental Results}
\subsubsection{Exp.1 - Lane change motion planning to a target lane}
For each lane change human driving sample $\boldsymbol{H}^i$, the target lane is treated as a known value according to $\boldsymbol{T}_{GT}^i$, hence a set of trajectories ${\cal T}(\boldsymbol{S}^i)$ is synthesized to the target lane according to the driving situation $\boldsymbol{S}^i$.
 Tessellation of the terminal state is conducted on duration $\tau$ in range $[\tau_{min}=6s,\tau_{max}=10s]$ at resolution $1s$, and longitudinal velocity $\dot{s}$ in Eqn.~(\ref{sampledetails}) with $\Delta V=4m/s$ at resolution $1m/s$.
Since there's no lane selection decision in this experiment, only the total cost with traditional cost terms (refer to Eqn.~(\ref{total_cost_0})) is trained using the proposed approach in \ref{problem_formulation}.
The following measurements are defined to evaluate its performance:
\begin{eqnarray}
\label{eqn:dismeasure}
 MinDist^i &=& \min_j d_j^i \\
 MinCost^i &=& d_{j^*}^i
\end{eqnarray}
where $d_j^i = d(\boldsymbol{T}_j^i, \boldsymbol{T}_{GT}^i)$,  $j^* = \argmin_{j} f({\boldsymbol{T}}_j^i|\boldsymbol{S}^i,\boldsymbol{\omega})$. According to the definition, $MinDist^i$ is the minimal distance between the human driving trajectory and generated trajectories, representing the minimal error that the algorithm could achieve on sample $i$. $MinCost^i$ is the distance between the selected trajectory using the learned $f$ and the human driving trajectory, representing the actual error of the algorithm on sample $i$.

\begin{figure}[tb]
  \begin{center}
    \includegraphics[keepaspectratio=true,width=1\linewidth]{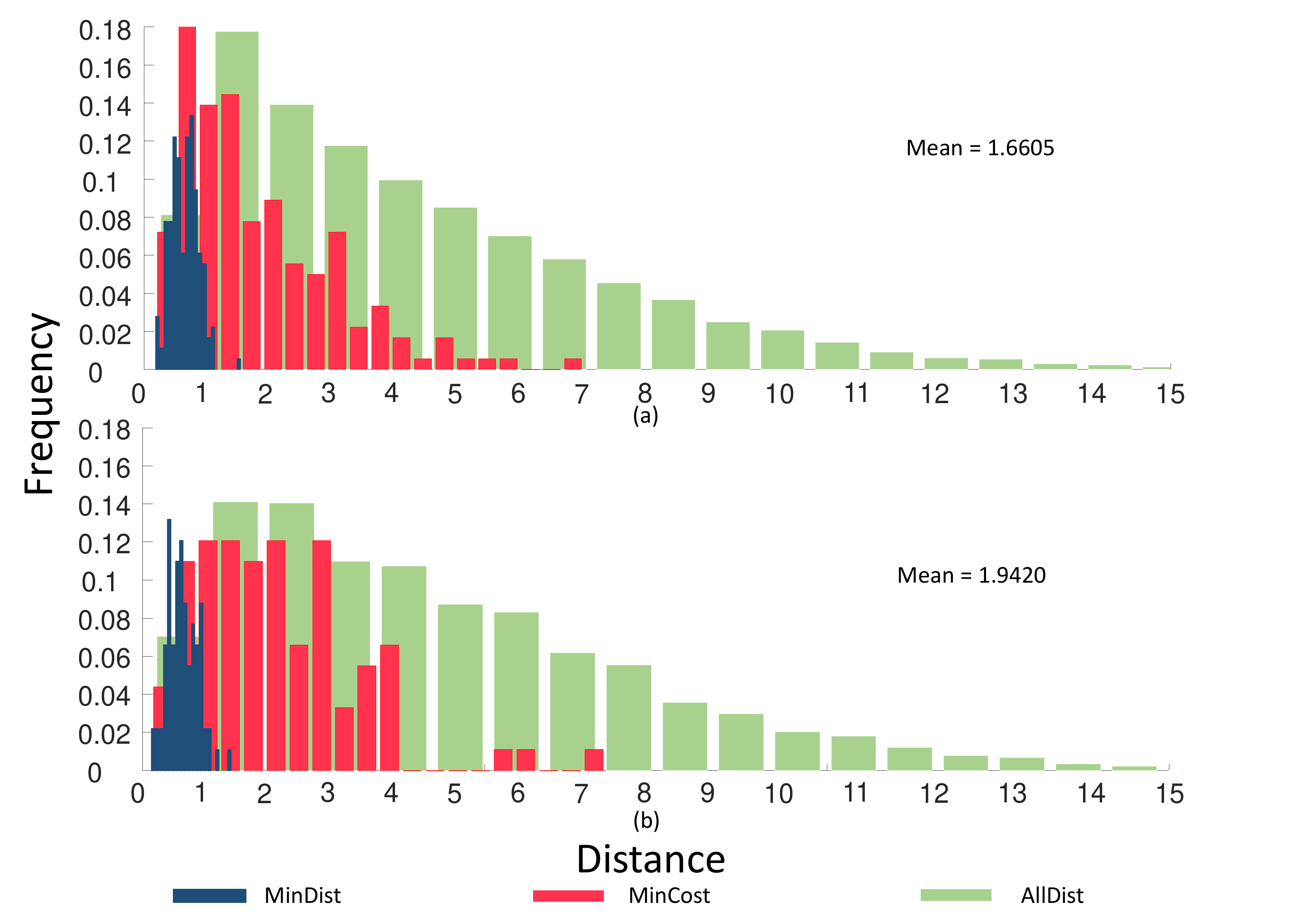}
  \end{center}
\caption{Result of Exp.1. Top: training. Bottom: testing.}
\label{Exp1_Train_Test}
\end{figure}

Histograms of $MinDist^i$ and $MinCost^i$ are generated describing the distance profiles of the most similar trajectory with the human driving one and the selected one by using the learnt cost function, in addition with a histogram of all $d_j^i$, denoted by ``$AllDist$''. These histograms are plotted in Fig.\ref{Exp1_Train_Test} in blue, red and green respectively. The mean value of $MinCost^i$ is also presented in the figure, which represents the average distance between planned trajectory and human-driving trajectory. From the results of both training and testing, it can be found that distribution of $MinDist^i$ is more concentrated to $0$ compared with $MinCost^i$, which means that the planner cannot always select the trajectory closest to the human-driving trajectory based on the learned cost. However, compared with distance distribution of all candidate trajectories (refer to histogram of $AllDist$), distribution of $MinCost^i$ is significantly closer to $0$, which means compared with random selection, the planner tends to select trajectory similar to human-driving ones under the guidance of learned cost, demonstrating effectiveness of the proposed method.

\subsubsection{Exp.2 - Lane change motion planning with simultaneous decision of the target lane}

\begin{figure}[tb]
  \begin{center}
    \includegraphics[keepaspectratio=true,width=1\linewidth]{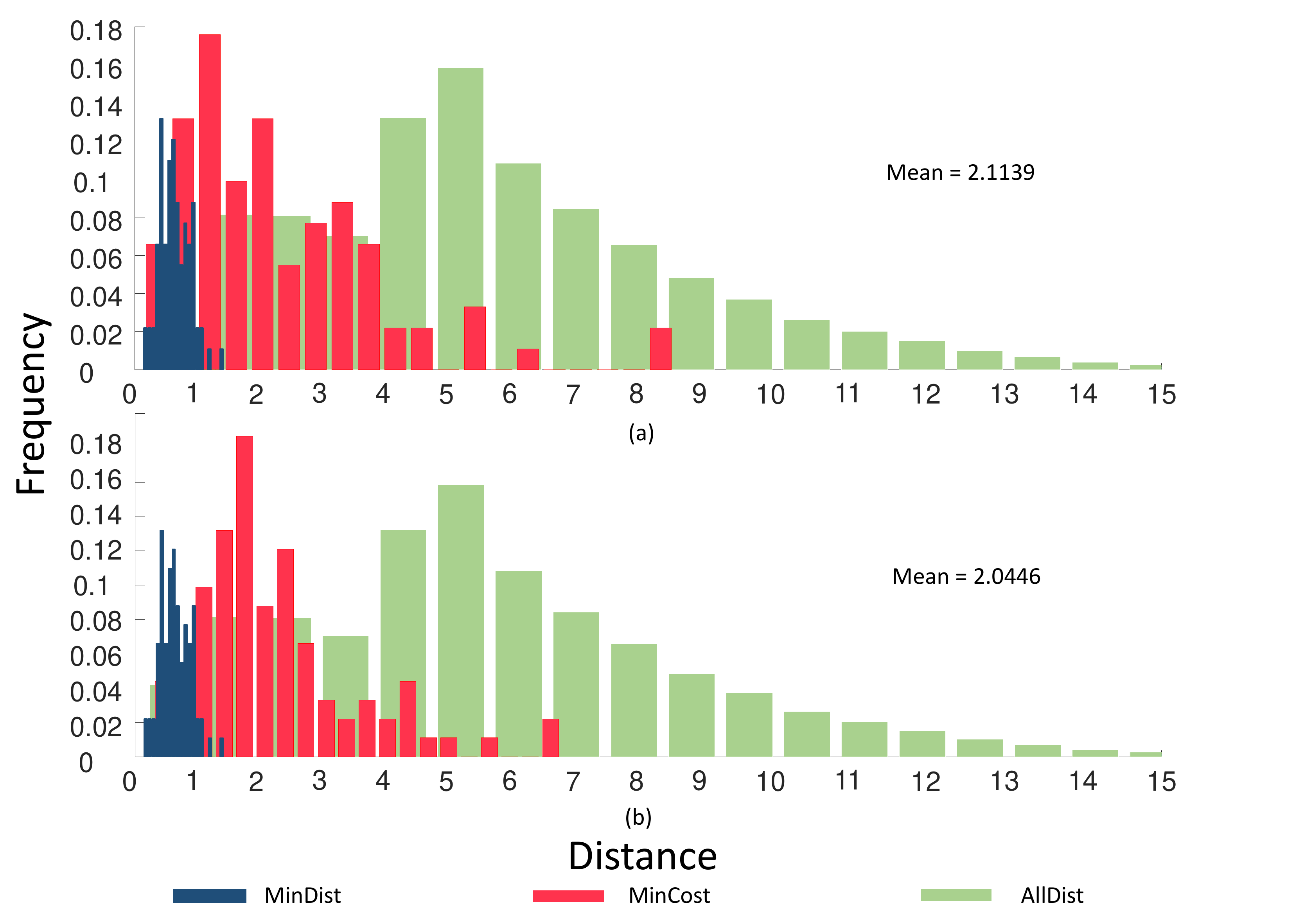}
  \end{center}
\caption{Result of Exp.2. Top: $f_0$. Bottom: $f_1$.}
\label{Exp2_Test}
\end{figure}

In this experiment, for each lane change human driving sample $\boldsymbol{H}^i$, ${\cal T}(\boldsymbol{S}^i)$ is generated to include lane change trajectories to both left and right lanes. Note that under this setting, the heuristic lane incentive cost can be introduced. Testing results produced by two versions of total cost ($f_0$ and $f_1$) are presented in Fig.~\ref{Exp2_Test}, which can be discussed and concluded in the same way with those in Exp. 1. Compared with results of Exp. 1 (Fig.~\ref{Exp1_Train_Test}), it can be noticed that the histogram of $AllDist$ moves obviously away from $0$, which is because candidate trajectories are generated to two target lanes and distance errors of generated trajectories whose targets lane are opposite to ground truth trajectories are taken into account. The fact implies that the problem is more challenging compared with Exp. 1. Due to the fact, the mean distance error of $f_0$ based model on testing set increases from Exp. 1 to Exp. 2: 1.9420 in Exp. 1 versus 2.1139 in Exp. 2 (refer to Fig~\ref{Exp1_Train_Test}(b) and Fig~\ref{Exp2_Test}(a)). By introducing heuristic lane incentive cost, the mean distance error of $f_1$ based model is 2.0446 (refer to Fig~\ref{Exp2_Test}) in Exp. 2, lower than that of $f_0$ based model, demonstrating effectiveness of the heuristic lane incentive cost. 

\begin{table}[tb]
	\centering
	\caption{Exp.2 - Decision of the target lane. \protect\\(GT): Ground truth (P): Prediction}
	\begin{tabular}{|c|c|c|c|c|}
		\hline
		\multicolumn{4}{|c|}{$f_0$ based model}         \\ \hline
		& LLC(P)   & RLC(P)  & recall  \\ \hline
		LLC(GT)  & {\textbf {38}}  &7    &84.44\%    \\ \hline
        RLC(GT)  & 4 & {\textbf{41}}   &91.11\%    \\ \hline
        Precision  & 90.48\% &  85.42\%   &  OA: \textbf{87.78\%} \\ \hline
        \multicolumn{4}{|c|}{$f_1$ based model}         \\ \cline{1-4}
		& LLC(P)   & RLC(P)  & recall  \\ \hline
		LLC(GT)  & {\textbf{41}} & 4    &91.11\%    \\ \hline
        RLC(GT)  & 3& {\textbf{42}}   &93.33\%   \\ \hline
        Precision  &  93.18\% & 91.30\% & OA: \textbf{92.22\%} \\ \hline
	\end{tabular}
    \label{Exp2Decision}
\end{table}

Since this experiment is conducted with unknown target lanes, the planned trajectory suggests not only the vehicle's motion sequence in future seconds, but also the target lane, i.e., simultaneous maneuver decision of either left or right lane changes.
Testing results based on $f_0$ and $f_1$ are analyzed on decision making aspect as summarized in Tab.~\ref{Exp2Decision} respectively.
Treating human driver's decision as the ground truth, the deduced behavioral decision of planned trajectories (refer to \ref{subsec_planning_and_decision}) is evaluated using precision and recall of each behavior type (LLC and RLC in Exp. 2) and overall accuracy (OA) is also calculated for comparison.
From the table, $f_1$ based model achieves higher recall and precision on both behaviors than $f_0$ based model, demonstrating that heuristic lane incentive cost introduced in $f_1$ helps to make human-like decision.

\subsubsection{Exp.3 - Motion planning with simultaneous decision of maneuver}

\begin{figure}[tb]
  \begin{center}
    \includegraphics[keepaspectratio=true,width=1\linewidth]{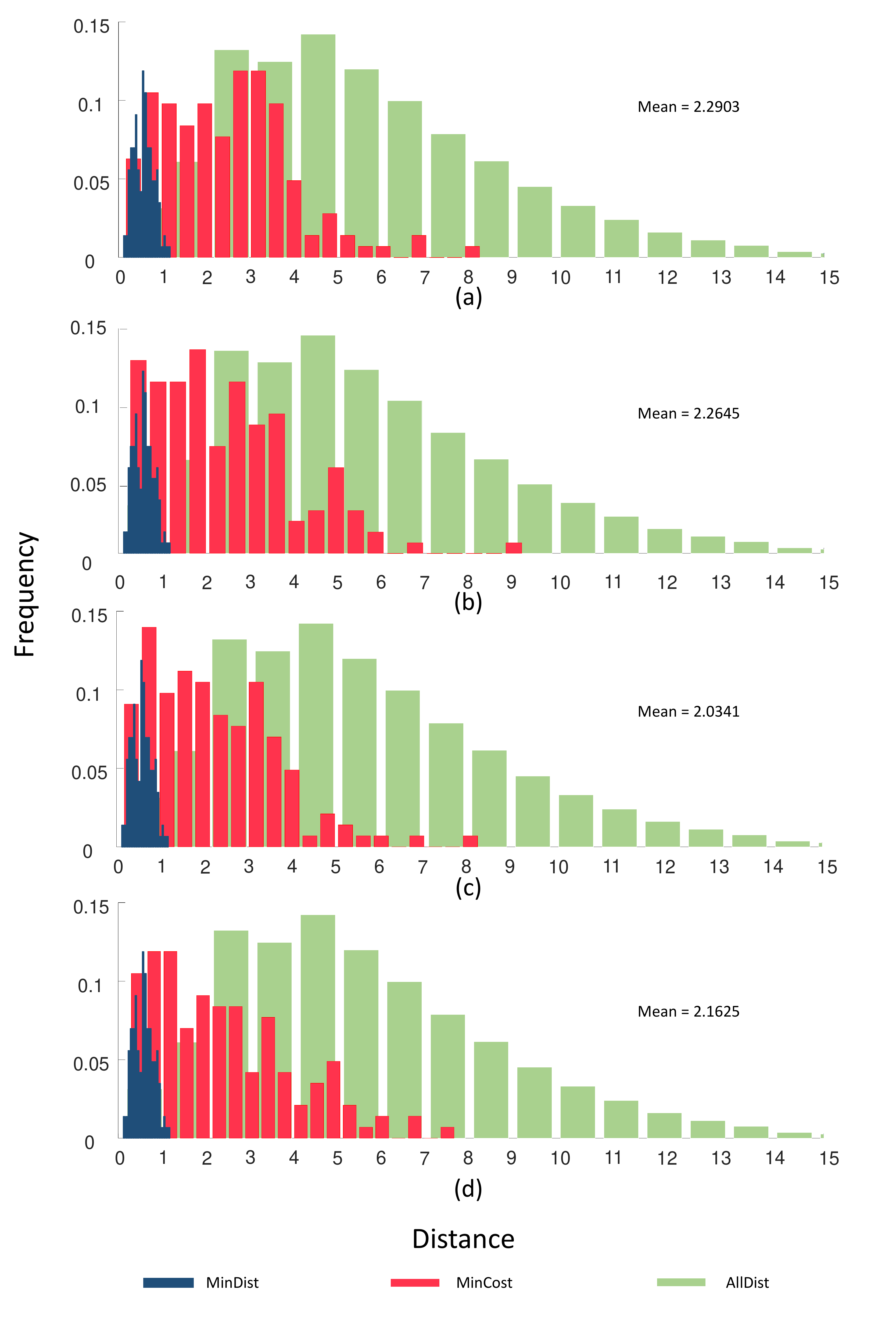}
  \end{center}
\caption{Result of Exp.3. Top to bottom: $f_0$, $f_1$, $f_2$ and $f_3$.}
\label{Exp3_Test}
\end{figure}

In this experiment, for each human driving sample $\boldsymbol{H}^i$, ${\cal T}(\boldsymbol{S}^i)$ is generated to include trajectories corresponding to behaviors of left lane changing, car following and right lane changing. Note that under this setting, the two versions of learning based lane incentive cost can be introduced. Testing results produced by four versions of total cost ($f_0$, $f_1$, $f_2$ and $f_3$) are presented in Fig.~\ref{Exp3_Test} that can be discussed in the same way, while it can be found that the histograms of ``MinCost'' have less focused picks and distribute across broader range comparing with the results of previous experiments.

The trajectory candidate of the highest probability is selected, which prompts a maneuver of LLC, RLC or CF too.
Testing results of maneuver decision based on  $f_0$, $f_1$, $f_2$ and $f_3$ are shown in Tab.~\ref{Exp3Decision}. To better understand how the learning based cost helps to improve the final decision performance, the decision results of $rf_2$ and $rf_3$ are also presented (refer to Tab.~\ref{RFDecision}). Comparing to the results of Exp. 2, the accuracies of $f_0$ and $f_1$ are much lower, indicating that the three-way lane selective decision prediction is much more challenging. Meanwhile, it can be observed that with the help of learning based lane incentive cost, models based on $f_2$ and $f_3$ perform considerably better. 

What's more, there are two points worthy of notice. 1) The overall accuracy of behavior decision by integrated costs ($f_2$ and $f_3$) are better than both results by $f_0$ and results of $rf_2$ and $rf_3$'s output, which indicates that the traditional costs and learning based lane incentive costs are complementary. 2) The overall accuracy of behavior decision by $f_2$ is better than that of $f_3$. From results of Exp. 2 (Tab.~\ref{Exp2Decision}), one can see that the traditional costs does well in predicting the lane change target, i.e., LLC and RLC will not be heavily confused by traditional cost based model, and a three-way random forest doesn't show any advantage or difference in distinguishing future behavior of LLC and RLC (refer to results of $rf_3$ in Tab.~\ref{Exp3Decision}). The interesting point is that their performance regarding car following behavior is quite different: traditional cost based model tends to predict car following behavior as lane change (refer to $f_0$ based model results in Tab.~\ref{Exp3Decision}) while random forest models tends to predict lane change behavior as car following (refer to Tab.~\ref{RFDecision}). In other words, the complementation between traditional cost and random forest takes place in respect of predicting maneuver rather than predicting lane change target. This may implies one of the reasons why $f_2$ based model performs better than $f_3$ based model: because the $f_2$ integrates lane incentive cost based on $rf_2$ which focuses on distinguishing whether lane change will happen while $f_3$ is based on $rf_3$ which also takes care of lane change target decision which actually cannot help traditional cost a lot.

\begin{table}[tb]
	\centering
	\caption{Exp.3 - Decision of the target lane. \protect\\(GT): Ground truth (P): Prediction}
	\begin{tabular}{|c|c|c|c|c|}
		\hline
		\multicolumn{5}{|c|}{$f_0$ based model}         \\ \hline
		&CF(P) & LLC(P)   & RLC(P)  & recall  \\ \hline
        CF(GT) & {\textbf {22}} & 15   & 16  &41.51\%     \\ \hline
		LLC(GT) & 3 & {\textbf {40}}  &2    &88.89\%    \\ \hline
        RLC(GT) & 5 & 2 & {\textbf{38}}   &84.44\%    \\ \hline
        Precision & 73.33\% & 70.18\% &  67.86\%   &  OA: \textbf{69.93\%} \\ \hline
        \multicolumn{5}{|c|}{$f_1$ based model}         \\ \cline{1-5}
		&CF(P) & LLC(P)   & RLC(P)  & recall  \\ \hline
        CF(GT)  & {\textbf{34}} & 11    & 8 &64.15\%     \\ \hline
		LLC(GT)  & 2 & {\textbf{38}} & 5    &84.44\%    \\ \hline
        RLC(GT) & 10 & 2& {\textbf{33}}   &73.33\%   \\ \hline
        Precision & 73.91\%  &  74.51\% & 71.74\% & OA: \textbf{73.43\%} \\ \hline
        \multicolumn{5}{|c|}{$f_2$ based model}         \\ \cline{1-5}
		&CF(P) & LLC(P)   & RLC(P)  & recall  \\ \hline
        CF(GT)  & {\textbf{44}} & 4    & 5 &83.02\%     \\ \hline
		LLC(GT)  & 0 & {\textbf{38}} & 7    &84.44\%    \\ \hline
        RLC(GT) & 3 & 1& {\textbf{41}}   &91.11\%   \\ \hline
        Precision & 93.62\%  &  88.37\% & 77.36\% & OA: \textbf{86.01\%} \\ \hline
        \multicolumn{5}{|c|}{$f_3$ based model}         \\ \cline{1-5}
		&CF(P) & LLC(P)   & RLC(P)  & recall  \\ \hline
        CF(GT)  & {\textbf{45}} & 4    & 4 &84.91\%     \\ \hline
		LLC(GT)  & 5 & {\textbf{37}} & 3    &82.22\%    \\ \hline
        RLC(GT) & 9 & 4& {\textbf{32}}   &71.11\%   \\ \hline
        Precision & 76.27\%  &  82.22\% & 82.05\% & OA: \textbf{79.72\%} \\ \hline
	\end{tabular}
    \label{Exp3Decision}
\end{table}

\begin{table}[tb]
	\centering
	\caption{Exp.3 - Decision of the target lane by $rf_2$ and $rf_3$. \protect\\(GT): Ground truth (P): Prediction}
	\begin{tabular}{|c|c|c|c|c|}
		\hline
		\multicolumn{5}{|c|}{result of $rf_2$}         \\ \hline
		&CF(P) & \multicolumn{2}{|c|}{LC(P)}   & recall  \\ \hline
        CF(GT) & {\textbf {46}} & \multicolumn{2}{|c|}{7}  &86.79\%     \\ \hline
		LC(GT) & 26 & \multicolumn{2}{|c|}{\textbf {64}}     &71.11\%    \\ \hline
        Precision & 63.89\% & \multicolumn{2}{|c|}{90.14\%}   &  OA: \textbf{76.92\%} \\ \hline
        \multicolumn{5}{|c|}{result of $rf_3$}         \\ \cline{1-5}
		&CF(P) & LLC(P)   & RLC(P)  & recall  \\ \hline
        CF(GT)  & {\textbf{49}} & 2    & 2 &92.45\%     \\ \hline
		LLC(GT)  & 11 & {\textbf{29}} & 5    &64.44\%    \\ \hline
        RLC(GT) & 11 & 2& {\textbf{32}}   &71.11\%   \\ \hline
        Precision & 69.01\%  &  87.88\% & 82.05\% & OA: \textbf{76.92\%} \\ \hline
	\end{tabular}
    \label{RFDecision}
\end{table}

\subsection{Discussion}
\subsubsection{The hyper-parameter $K$}
The total cost (refer to \ref{subsec_total_cost}) is defined as weighted sum of $k=1,2,\cdots,K$ power of various cost terms with weighting parameters $\boldsymbol{\omega}$ to be learned from data. We train the model with traditional cost (refer to Eqn.~\ref{total_cost_0}) with $K=1,\cdots,5$ and experimental setting of Exp. 3 to see how training loss as well as overall accuracy (OA) of target lane decision on training set varies with different $K$. The results are shown in Tab.~\ref{exp_various_K}. From the results we see that increasing $K$ helps achieve smaller loss and higher overall accuracy of decision, meaning that the model can fit training data better. However, a larger $K$ could face the risk of overfitting. In this research, we set $K=5$ heuristically. In future work, elaborating this parameter setting should be addressed.  
\begin{table}[tb]
	\centering
	\caption{Training results of $f_0$ based model with various $K$}
	\begin{tabular}{|c|c|c|c|c|c|}
		\hline
		& 1 & 2 & 3 & 4 & 5  \\ \hline
		Loss & 639.44 & 591.67 & 558.20 & 551.33 & 544.30 \\ \hline
        Decision OA & 62.69\% & 65.19\% & 68.15\% & 77.78\% & 79.26\% \\ \hline
	\end{tabular}
	\label{exp_various_K}
\end{table}

\subsubsection{The unbalanced driving behavior problem}
Car following is the most dominant behavior in daily driving, while lane change is an occasional one.
In order to simulate such a situation, an experiment is conducted by using a large proportion of car following samples.
The experimental design follows Exp.3, the cost function is $f_0$, and the results are shown in Tab.~\ref{exp_more_CF}.
From the table, we
see that the overall accuracy is higher than that in Tab.~\ref{Exp3Decision}.
However, this is at the cost of severe decrease of recall of
lane change decisions, i.e., the model tends to predict a lane
change decision as car following, which is not actually desired
in real-world application.
It is therefore important to balance the number of samples of different behaviors in dataset as what we did in the experiment,
and prior distribution of different behaviors could be incorporated in online prediction.
\begin{table}[tb]
	\centering
	\caption{A comparative experiment with Exp. 3 using large proportion of car following samples\protect\\(GT): Ground truth (P): Prediction}
	\begin{tabular}{|c|c|c|c|c|}
		\hline
		\multicolumn{5}{|c|}{result on training set}         \\ \hline
		&CF(P) & LLC(P)   & RLC(P)  & recall  \\ \hline
        CF(GT)  & {\textbf{783}} & 9    & 18 &96.67\%     \\ \hline
		LLC(GT)  & 26 & {\textbf{60}} & 4    &66.67\%    \\ \hline
        RLC(GT) & 16 & 2& {\textbf{74}}   &82.22\%   \\ \hline
        Precision & 94.91\%  &  86.96\% & 77.08\% & OA: \textbf{92.63\%} \\ \hline
        \multicolumn{5}{|c|}{result on testing set}         \\ \cline{1-5}
		&CF(P) & LLC(P)   & RLC(P)  & recall  \\ \hline
        CF(GT)  & {\textbf{390}} & 0    & 15 &96.30\%     \\ \hline
		LLC(GT)  & 17 & {\textbf{27}} & 1    &60.00\%    \\ \hline
        RLC(GT) & 18 & 0& {\textbf{27}}   &60.00\%   \\ \hline
        Precision & 91.76\%  &  100.00\% & 62.79\% & OA: \textbf{89.70\%} \\ \hline
	\end{tabular}
    \label{exp_more_CF}
\end{table}

\section{Conclusion and Future Work}

Aiming at human-like autonomous driving, this research proposes a motion planning method by learning from naturalistic data. 
A cost function is formulated by incorporating not only the components on trajectory's comfort, efficiency and safety, but also lane incentive by referring to a human driver's lane change decisions, where two version of lane incentive costs (heuristic and learning based) are proposed. 
A method is developed to learn cost coefficients by correlating the probability of a trajectory being selected with its distance to the human driving one at the same driving situation.
A data set is developed using the naturalistic human driving data on the motorways in Beijing, containing samples of lane changes to the left and right lanes, and car followings.
Experiments have been conducted on three aspects: 1) lane change motion planning to a given target lane; 2) lane change motion planning with simultaneous decision of a target lane; and 3) motion planning with simultaneous decision of maneuver.
Results show that the selected trajectory is among the closest to human drivers' in the set of all trajectories.
In addition, the proposed method allows simultaneous motion planning and behavior decision. 
In case lane change has been decided, with the heuristic lane incentive cost involved, decision accuracy to either the left and right lane is above 90\% treating human drivers' as the ground truth.
In case either lane change or car following/longitudinal driving are candidates, with the help of learning based lane incentive cost, the decision accuracy is above 86\%. The model can be used as a human-like trajectory planner for autonomous vehicles. It can also be treated as a driving behavior model and used to build real-world traffic simulator or perform early prediction of human driver's lane selective intention.

The proposed method can be improved in future by elaborating the following issues: 1) Distance metric. The data of this paper were collected during non-rush hour, where the traffic was smooth, and the vehicles' headway distance were mostly more than ten meters. The dataset does not contain bus, truck or other types of large vehicles too. It is therefore the vehicles' shape, size and heading are not fully addressed in the distance metric and safety cost of this paper. However, at the scenarios where various types of vehicles exist or close interactions happen, vehicle shape and heading are crucial factors that need to be elaborated in future. Furthermore, high-level features can be incorporated too to weight the impacts of features in different behaviors. 2) Interactive behaviors during lane change. In our early work \cite{48}, more than 1000 naturalistic lane change samples were analyzed. It was found that 72\% of the samples are self-motivated, have no significant interaction with the surrounding vehicles, and the environmental vehicles keep their initial states such as speed and lane during the whole procedure. This paper has been focused on trajectory planning at such scenarios. Hence a linear prediction model for environmental vehicles can be reasonably adopted in evaluating safety cost, and we directly encourage the complete planned trajectory to approximate the human-driving one for training and compare their distance at testing phase. In the future, the interactive behaviors will be studied by considering more complex prediction model and introducing re-planning during lane change procedure.

\ifCLASSOPTIONcaptionsoff
  \newpage
\fi

\bibliographystyle{IEEEtran}
\bibliography{IEEEabrv,ref}

\begin{IEEEbiography}[{\includegraphics[width=1in,height=1.25in,clip,keepaspectratio]{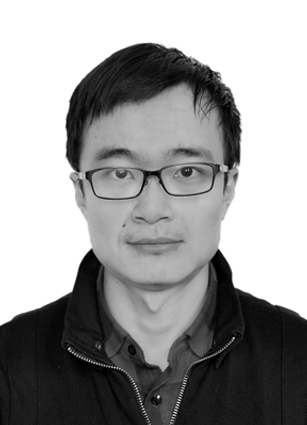}}]{Donghao Xu}
received B.S. degree in information and computing science in 2012
from Peking University, China. In 2018, he obtained Ph.D. degree in computer science from the same university. He is currently a post-doctoral researcher in computer science in the Key Lab of Machine Perception (MOE), Peking University, China. His research interests include computer vision, machine learning and intelligent vehicles.
\end{IEEEbiography}

\begin{IEEEbiography}[{\includegraphics[width=1in,height=1.25in,clip,keepaspectratio]{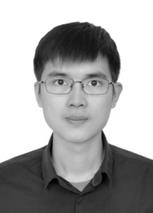}}]{Zhezhang Ding}
Zhezhang Ding received the B.S. degree in computer science (intelligent science and technology) from Peking University, Beijing, China, in 2018. He is currently pursuing the M.S. degree in intelligent robots with the Key Laboratory of Machine Perception, Peking University, Beijing, China 
His research interests include intelligent vehicles and machine learning.
\end{IEEEbiography}

\begin{IEEEbiography}[{\includegraphics[width=1in,height=1.25in,clip,keepaspectratio]{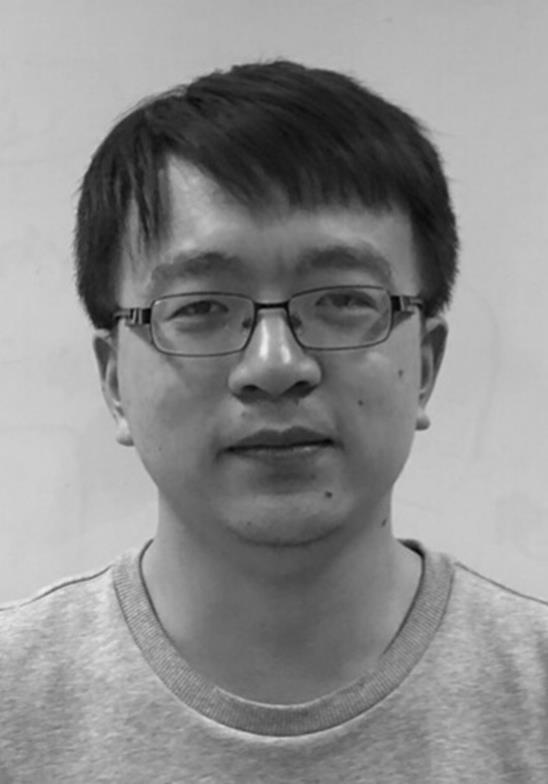}}]{Xu He}
Xu He received the B.S. degree in automation from Tsinghua University, Beijing, China, in 2015 and the M.S. degree in intelligent science and technology from Peking University, Beijing, in 2018. He is now working in China Aerospace Science and Industry Corp. His research interests include intelligent vehicle, driving behavior learning and motion planning.
\end{IEEEbiography}

\begin{IEEEbiography}[{\includegraphics[width=1in,height=1.25in,clip,keepaspectratio]{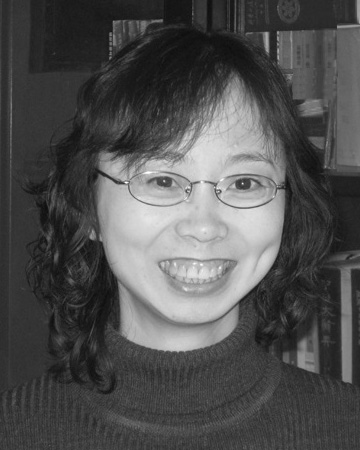}}]{Huijing Zhao}
received B.S. degree in computer science in 1991 from
Peking University, China. From 1991 to 1994, she was recruited by Peking
University in a project of developing a GIS platform. She obtained M.E.
degree in 1996 and Ph.D. degree in 1999 in civil engineering from the
University of Tokyo, Japan. After post-doctoral research
as the same university, in 2003, she was promoted to be a visiting
associate professor in Center for Spatial Information Science, the
University of Tokyo, Japan. In 2007, she joined Peking Univ as an associate
professor at the School of Electronics Engineering and Computer Science.
Her research interest covers intelligent vehicle, machine perception and mobile robot.
\end{IEEEbiography}

\begin{IEEEbiography}[{\includegraphics[width=1in,height=1.25in,clip,keepaspectratio]{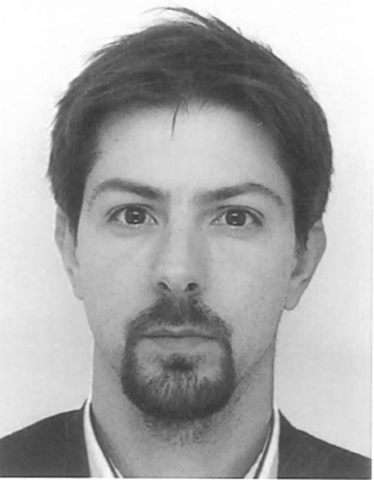}}]{Mathieu Moze}
was born in Bordeaux, France, in 1979. He received a M.Eng. degree in Mechatronics from French Ecole Nationale d'Ingenieurs (ENIT) and a M.S. degree in Control Systems from Institut National Polytechnique Toulouse (INPT),  both  in  2003, before a Ph.D. degree in Control Theory from Bordeaux University in 2007.
In 2008, he became a consultant for industrial firms and was associated with IMS Laboratory in Bordeaux, where he studied algebraic approaches to fractional order systems analysis and robust control theory. 
Since 2010, he  has been with the Scientific Department of Groupe PSA where he conducts advanced research concerning modeling, design and control of mechatronic systems for automotive applications, mainly Autonomous Driving.

\end{IEEEbiography}

\begin{IEEEbiography}[{\includegraphics[width=1in,height=1.25in,clip,keepaspectratio]{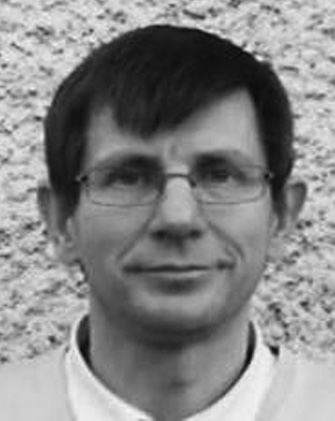}}]{Fran\c{c}ois Aioun}
received an engineering degree in electronics, computer science and Automatic control in 1988 from
ESIEA high school, France. In 1989, he obtained a post-graduate diploma in Automatic control and Signal processing.
From 1989 to 1993, he was recruited by Electricit\'e de France to study active vibration control of a structure.
He obtained a Ph.D. degree in Automatic Control in 1993.
After post-doctoral research at Ecole Normale sup\'erieure of Cachan and in different companies, he joined Groupe PSA in 1997.
His research interest in automatic control covers active vibration, power plant, powertrain, actuators and more recently Autonomous
and Intelligent vehicles.
\end{IEEEbiography}

\begin{IEEEbiography}[{\includegraphics[width=1in,height=1.25in,clip,keepaspectratio]{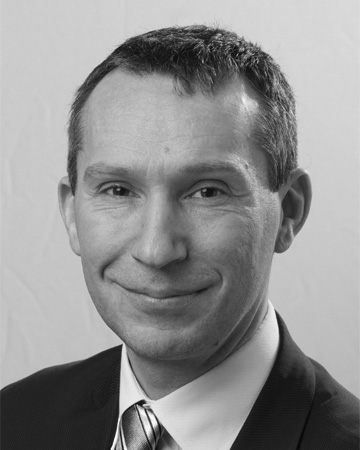}}]{Franck Guillemard}
was born in France in March 18th, 1968. He obtained his Ph.D. degree in Control Engineering from the University of Lille, France, in 1996. Presently he works in the Scientific Department of Groupe PSA where he is in charge of advanced research concerning Computing Science, Electronics, Photonics and Control. He is also expert in the field of modeling, design and control of mechatronic systems for automotive.
\end{IEEEbiography}

\end{document}